%% file: aistat2015.tex
\documentclass[twoside]{article}

\usepackage{amsmath}
\usepackage{amsthm}
\usepackage{amssymb}
\usepackage{algorithm}
\usepackage{algorithmic}
\usepackage{enumitem}
\usepackage{hyperref}
\usepackage{amsfonts}
\usepackage{url}
\usepackage{defs}
\usepackage{graphicx}
\usepackage{caption}
\usepackage{subcaption}
\usepackage{breqn}
\usepackage{booktabs}

\newcommand{\eat}[1]{}
\usepackage[accepted]{aistats2016}
\begin{document}

%

%

\twocolumn[

\aistatstitle{Scalable MCMC for Mixed Membership\\Stochastic Blockmodels}

\aistatsauthor{ Wenzhe Li* \And Sungjin Ahn* \And Max Welling }

\aistatsaddress{University of Southern California \And   University of Montreal \And University of Amsterdam } ]

\begin{abstract}
We propose a stochastic gradient Markov chain Monte Carlo (SG-MCMC) algorithm for scalable inference in mixed-membership stochastic blockmodels (MMSB). Our algorithm is based on the stochastic gradient Riemannian Langevin sampler and achieves both faster speed and higher accuracy \emph{at every iteration} than the current state-of-the-art algorithm based on stochastic variational inference. In addition we develop an approximation that can handle models that entertain a very large number of communities. The experimental results show that SG-MCMC strictly dominates competing algorithms in all cases.
\end{abstract}

\section{Introduction}

Probabilistic graphical models represent a convenient paradigm for modeling complex relationships between a potentially very large number of random variables. Bayesian graphical models [13], where we define priors and infer posteriors over parameters also allow us to quantify model uncertainty and facilitate model selection and averaging. But an increasingly urgent question is whether these models and their inference procedures will be up to the challenge of handling very large ``big data'' problems.

A large subclass of Bayesian graphical models is represented by so called ``topic models'' such as latent Dirichlet allocation [4]. For these type of models very efficient inference algorithms have recently been developed, either based on stochastic variational Bayesian inference (SVB) [7,12] or on stochastic gradient Markov chain Monte Carlo (SG-MCMC) [2,3,5,6,19]. Both methods have the important property that they only require a small subset of the data-items for every iteration. In other words, they can be applied to (infinite) streaming data.

An important class of ``big data'' problems are given by networks. Large networks such as social networks easily run into billions of edges and tens of millions of nodes. An interesting problem in this area is the discovery of communities: densely connected groups of nodes that are only sparsely connected to the rest of the network. Large networks may contain millions of such communities. To model overlapping communities the mixed membership stochastic blockmodel (MMSB) was introduced in [8]. Very recently, an efficient stochastic variational inference algorithm was developed for a special case, the assortative MMSB (a-MMSB) [1], greatly extending the reach of Bayesian posterior inference to realistic large scale problem settings. Inspired by this work, and earlier comparisons between SVB and SG-MCMC on LDA [2] we developed a scalable SG-MCMC algorithm for a-MMSB and compared it against SVB on the community detection problem.

Our conclusion is consistent with the findings of [2], namely that SG-MCMC is also both faster and more accurate than SVB algorithms in this domain. While one should expect SG-MCMC to be more accurate than SVB asymptotically (SVB is asymptotically biased while SG-MCMC is not), it is interesting to observe that SG-MCMC dominates SVB across all iterations, despite the fact that SG-MCMC should have a larger variance contribution to the error.

\section{Assortative Mixed-Membership Stochastic Blockmodels}
Assortative mixed-membership stochastic blockmodel (a-MMSB) [1] is a special case of MMSB [8] that models the group-structure in a network of $N$ nodes. In particular, each node $a$ in the node set $\cV^*$ has a $K$-dimension probability distribution $\pi_a$ of participating in the $K$ members of the community set $\cK$. For every possible peer $b$ in the network, each node $a$ randomly draws a community $z_{ab}$. If a pair of nodes ($a,b$) in the edge set $\cE^*$ are in the same community: $z_{ab}=z_{ba} = k$, then they have a significant probability $\bt_k$ to connect, i.e., $y_{ab}=1$. Otherwise this probability is small. Each community has its connection strength $\beta_{k} \in (0,1)$ which explains how likely its members are linked to each other.

The generative process of a-MMSB is then given by,

\benum[topsep=-0pt,itemsep=-0ex,partopsep=1ex,parsep=1ex]
\item For each community $k$, draw community strength $\bt_k \sim \mbox{Beta}(\eta)$
\item For each node $a$, draw community memberships $\pi_a \sim \mbox{Dirichlet}(\al)$
\item For each pair of nodes $a$ and $b$,
\benum[topsep=-0pt,itemsep=-0ex,partopsep=1ex,parsep=1ex]
	\item Draw interaction indicator $z_{ab} \sim \pi_a$
    \item Draw interaction indicator $z_{ba} \sim \pi_b$
    \item Draw link $y_{ab} \sim \mbox{Bernoulli}(r)$, where $r = \bt_k$ if $z_{ab}=z_{ba}=k$, and $r=\dt$ otherwise.
\eenum
\eenum

Unlike the a-MMSB, the original MMSB maintains pair-wise community strength $\bt_{k,k'}$ for all pairs of the communities. Note that it is trivial to extend the results that we obtain in this paper to the general MMSB model. The joint probability of the above process can be written as:
\begin{align}
p(y,z,\pi ,\bt | \al, \eta) &= \prod_{a=1}^{N} \prod_{b > a}^{N} p(y_{ab} | z_{ab}, z_{ba}, \bt) p(z_{ab}|\pi_a)\nonumber \\ &p(z_{ba}|\pi_b) \pd{a}{N}p(\pi_a|\al) \pd{k}{K} p(\bt_k|\eta).\label{eqn:joint}
\end{align}

Both variational inference [1,4,16] and collapsed Gibbs sampling algorithms [11] have been used successfully for small to medium scale problems. However, the $\cO(N^2)$ computational complexity per update prevents it from being applied to large scale networks. A stochastic variational algorithm was developed in [1] to address this issue, where each update only depends on a small mini-batch of the nodes in the network.

\section{Stochastic Gradient MCMC Algorithms}
Our algorithm will be based on the stochastic gradient Langevin dynamics (SGLD) [3]. To sample from a posterior distribution $p(\ta|\cX) \propto p(\cX|\ta) p(\ta)$ given $N$ i.i.d. data points $\cX=\{x_i\}_{i=1}^{N}$, SGLD applies the following update rule:
\bea
\ta^* \law \ta + \f{\epsilon_t}{2}\left(\nabla_{\ta}\log p(\ta_t)+N\bar{g}(\ta;\cD_n)\right) + \xi, \label{eqn:sgld_update}
\eea
where $\xi \sim \cN(0,\epsilon_t )$ with $\ep_t$ the step size, $\cD_n$ a mini-batch of size $n$ sampled from $\cX$, and $\bar{g}(\ta;\cD_n) = \frac{1}{|\cD_n|}\sum_{x\in \cD_n}^{}\grad_{\ta} \log p(x|\ta)$. As the step size goes to zero by a schedule satisfying $\sm{t}{\infty}\ep_t = \infty$ and $\sm{t}{\infty}\ep_t^2 < \infty$, SGLD samples from the true posterior distribution. In SGLD, the Metropolis-Hastings (MH) accept-reject tests are ignored since the rejection probability goes to zero as the step size collapses to zero. While for a finite step size this results in some bias, the overall error is reduced by the reduction of variance due to the ability to draw many more samples per unit time.

SGLD originated from the Langevin Monte Carlo (LMC) [15] where, unlike SGLD, the gradient is computed exactly using all data points and then a Metropolis-Hastings accept-reject test is applied. Because at each iteration SGLD requires to process only a mini-batch $\cD_n$ and ignores the MH test, the computation complexity per iteration is only $\cO(n)$ as opposed to $\cO(N)$ of LMC. Any mini-batch sampling algorithm in the form of Eqn. \eqref{eqn:sgld_update} is called valid SGLD as long as it guarantees the gradient estimator to be unbiased, i.e.,
$\eE_{\cD_n} \left[N\bar{g}(\ta;\cD_n)\right] = \grad_{\ta} \log p(\cX|\ta)\label{eqn:sgld_grad_estm}$ and the variance to be finite [5].


The stochastic gradient Riemannian Langevin dynamics (SGRLD) [2] is a subclass of SGLD which is developed to sample from the probability simplex. By applying Riemannian geometry [15] and using the mini-batch estimator in Eqn. \ref{eqn:sgld_update}, it achieved the state-of-the-art performance for latent Dirichlet allocation (LDA). In particular, for a $K$-dimensional probability simplex $\pi$, it uses the \textit{expanded-mean} re-parameterization trick, where the probability of a category $k$ is given by $\pi_k = \ta_k / \sum_{j=1}^K \ta_j$ with $\ta_k \sim \text{Gamma}(\al,1)$ and $\al$ a hyperparameter of the Dirichlet distribution $p(\pi | \al )$.
Then, the update rule becomes


\bea
\ta_{k}^* \law \left| \ta_{k} + \f{\ep}{2} \left( \al - \ta_{k} + \f{N}{|\cD_n|} \sum_{d \in \cD_n} g_d(\ta_{k}) \right) + (\ta_{k})^\ha \xi_{} \right|. \label{eqn:sgrld_update}
\eea
here $g_d(\ta_{k})$ is the gradient of the log posterior w.r.t. $\ta_{k}$ on a data point $d\in \cD_n$.

\section{Scalable MCMC for a-MMSB}
Our algorithm iterates updating local parameters $\pi$ and a global parameter $\beta$. Because both parameters lie on the probability simplex, we start from the SGRLD and modify it to be more efficient. Also, we introduce parameters $\phi$ and $\ta$ to re-parameterize $\pi$ and $\beta$ respectively. Thus, we alternatingly sample in the $\phi$ and $\ta$ spaces, and then obtain $\pi$ and $\beta$ by normalizing $\phi$ and $\ta$. From Eqn. \ref{eqn:joint}, summing over the latent variable $z$, we obtain the following joint probability,
\bea
&p(y, \pi , \bt | \al , \eta) = \prod_{a} p(\pi_a | \al ) + \prod_{a} p(\bt_k | \eta)\nn \\
&+ \prod_{a} \prod_{b > a}\sum_{z_{ab}, z_{ba}} p(y_{ab}, z_{ab}, z_{ba} | \bt , \pi_a, \pi_b)
\label{eqn:log_joint}.
\eea

\subsection{Sampling the global parameter}

By the re-parameterization, we have $\beta_k=\ta_{k1}/(\ta_{k0}+\ta_{k1})$, where $\ta_{ki}\sim$ Gamma$(\eta)\propto \ta_{ki}^{\eta-1}e^{-\ta_{ki}}$. Because $p(y, \pi , \bt | \al , \eta)$ decomposes into $p(y,\bt|\pi,\eta) p(\pi|\al)$, replacing $\bt$ by $\ta$, we compute the derivative of log of Eqn. \ref{eqn:log_joint} w.r.t. $\ta_{ki}$ for $i=\{0,1\}$ as follows:
\bea
\fp{\ln p(y ,\ta | \pi, \eta )}{\ta_{ki}} =
\fp{}{\ta_{ki}} \ln p(\ta_{ki} | \eta) + \sum_{a}\sum_{b>a} g_{ab}(\ta_{ki}),
\label{eqn:grad_bt}
\eea
where $g_{ab}(\ta_{ki}) = \fp{}{\ta_{ki}} \ln \sum_{z_{ab},z_{ba}} p(y_{ab},z_{ab},z_{ba} | \ta , \pi_a, \pi_b)$ which, similar to SGRLD for LDA [2], we can rewrite as
\bea
g_{ab}(\ta_{ki}) = \eE \left[ \eI[z_{ab}=z_{ba}=k] \left( \f{|1-i-y_{ab}|}{\ta_{ki}} - \f{1}{\ta_{k}} \right)\right]. \label{eqn:G}
\eea
where $\ta_{k}=\sum_i \ta_{ki}$ and $\eI[S]$ is equal to $1$ if a condition $S$ is TRUE and $0$ otherwise. The expectation is w.r.t. the posterior distribution of latent variables $z_{ab}$ and $z_{ba}$,
\bea
&&p(z_{ab}=k,z_{ba}=l | y_{ab}, \pi_a, \pi_b, \bt )\\ &&\propto f_{ab}^{(y)}(k,l)=
\begin{cases}
\bt_k^y(1-\bt_k)^{(1-y)}\pi_{ak}\pi_{bk}, & \mbox{if } k=l\\
\dt^{y}(1-\dt)^{(1-y)} \pi_{ak} \pi_{bl}, & \mbox{if } k \neq l \nn
\end{cases}
\label{eqn:case}
\eea
here we used simple notation $y$ instead of $y_{ab}$. Unlike the SGRLD for LDA [2], we compute the expectation in Eqn. \eqref{eqn:G} analytically by computing the normalization constant $Z_{ab}^{(y)} = \sm{k}{K}\sm{l}{K} f_{ab}^{(y)}(k,l)$ which can be reduced to $\cO(K)$ computation as follows
\begin{align}
&Z_{ab}^{(y)} = \dt^{y}(1-\dt)^{(1-y)} \nonumber\\&+\sm{k}{K} \left( \bt_k^{y}(1-\bt_k)^{(1-y)} - \dt^{y}(1-\dt)^{(1-y)}\right)\pi_{ak}\pi_{bk}
\end{align}

Then Eqn. \eqref{eqn:G} becomes
\bea
g_{ab}(\ta_{ki})
&=& \f{f_{ab}^{(y)}(k,k)}{Z_{ab}^{(y)}} \left(\f{|1-i-y|}{\ta_{ki}} - \f{1}{\ta_k} \right).
\eea
Plugging this into Eqn. \ref{eqn:sgrld_update}, we obtain the update rule for the global parameter,

\bea
\ta_{ki}^* \law \left| \ta_{ki} + \f{\ep}{2} \left\{ \eta - \ta_{ki} + h(\cE_{n_t})\sum_{(a,b) \in \cE_{n_t}}g_{ab}(\ta_{ki})\right\} \right.\nn \\ \left.+ (\ta_{ki})^{\ha}\xi_{ki} \right|, \label{eqn:global_update}
\eea
here $\cE_{n_t}$ is a mini-batch of $n_t$ node pairs sampled from $\cE^*$ for which we use the following strategy.

\textbf{Stratified sampling:} considering that the number of links is much smaller than that of non-links, we can reduce the variance of the gradient using stratified sampling, similar to the method used in [1]. For this, at every iteration we first randomly select a node $a$ and then toss a coin with probability 0.5 to decide whether to sample link edges or non-link edges for node $a$. If it is a link, we assign all of the link edges of node $a$ to $\cE_{n_t}$. Otherwise, i.e. if it is non-link, we uniformly sample a mini-batch of $N/m$ non-link edges from the entire set of non-link edges and assign it to $\cE_{n_t}$. Here, the $m$ is a hyper-parameter. Note that the size of $|\cE_{n_t}|$ will thus be much smaller than the total number of $N(N-1)/2$ edges when $m$ is reasonably large. Then, to ensure that the gradient is unbiased, a \textit{scaling parameter} $h(\cE_{n_t})$ is multiplied. Specifically, $h(\cE_{n_t})$ is set to $N$ when $\cE_{n_t}$ is a set of link edges and to $mN$ otherwise.

Because the global parameters $\{\beta_k\}$ does not change very fast compared to the local parameters $\{\pi_{ak}\}$, in practice we update only a random subset of the $\{\beta_k\}$ at each iteration.

\subsection{Sampling the local parameters}

Similar to the global parameter, we re-parameterize the local parameter $\pi_a$ such that $\pi_{ak} = \phi_{ak} / \sum_{j=1}^{K}\phi_{aj}$, with $\phi_{ak}\sim$ Gamma$(\alpha)\propto \phi_{ak}^{\alpha-1}e^{-\phi_{ak}}$. Then, taking the derivative of the log of Eqn. \ref{eqn:log_joint} w.r.t. $\phi_{ak}$, we obtain
\bea
\fp{\ln p(y , \phi | \bt, \al)}{\phi_{ak}} = \fp{}{\phi_{ak}} \ln p(\phi_{ak} | \al) +
\sum_{b} g_{ab}(\phi_{ak})
\eea
where $g_{ab}(\phi_{ak}) = \fp{}{\phi_{ak}} \ln \sum_{z_{ab}, z_{ba}} p(y_{ab}, z_{ab}, z_{ba} | \bt , \phi_a, \phi_b)$ which can be written as
\bea
g_{ab}(\phi_{ak}) = \eE\left[ \f{\eI [z_{ab} = k]}{ \phi_{ak}} - \f{1}{\phi_{a\cdot}} \right].
\label{eqn:grad_local}
\eea
Here the expectation is w.r.t. the distribution in Eqn. \eqref{eqn:case}. To compute the expectation analytically, we first integrate out $z_{ba}$ from Eqn. \eqref{eqn:case} because the expectation depends only on $z_{ab}$, and obtain the following probability up to a normalization constant

\begin{align}
&f_{ab}^{(y)}(k) = \sum_{l=1}^K f_{ab}^{(y)}(k,l)\nonumber \\ =& \pi_{ak}
\left\{ \bt_k^{y}(1-\bt_k)^{(1-y)} \pi_{bk} + \dt^{y}(1-\dt)^{(1-y)} (1- \pi_{bk}) \right\}.\label{eqn:local_f}
\end{align}

Then we obtain the normalization term by $Z_{ab}^{(y)} = \sm{k}{K}f_{ab}^{(y)}(k)$.
Integrating out the expectation in Eqn. \eqref{eqn:grad_local}, we obtain
\bea
g_{ab}(\phi_{ak}) = \f{f_{ab}^{(y)}(k)}{Z_{ab}^{(y)}\phi_{ak}} - \f{1}{\phi_a}.
\eea
Plugging this to Eqn. \ref{eqn:sgrld_update}, we obtain the SGRLD update rule for the local parameter $\phi_{ak}$
\bea
\phi_{ak}^* \law \left| \phi_{ak} + \f{\ep}{2} \left( \al - \phi_{ak} + \f{N}{|\cV_n|} \sum_{b \in \cV_n} g_{ab}(\phi_{ak})\right) \right. \nn \\ \left. + (\phi_{ak})^{\ha} \xi_{ak}\right|.\label{eqn:local_update}
\eea
Here, the $\cV_n$ is a random mini-batch of $n$ nodes sampled from $\cV^*$. Note that $|\cV_n| \ll |\cV^*|=N$.

\subsection{Scalable local updates for a large number of communities}

In some applications, the number of communities can be very large [18] so that the local update becomes very inefficient due to its $\cO(K|\cV_n|)$ computation per node in $\cE_{n_t}$ and also $\cO(KN)$ space complexity. In this section, we extend the above algorithm further with a novel approximation in order to make the algorithm scalable in terms of both speed and memory usage even for a very large number of communities which the SVI [1] approach cannot achieve.

\textbf{Community split:} for each node $a \in \cV^*$, we first split the community set $\cK$ into three mutually exclusive subsets: the \textit{active }set $\cA(a)$, the \textit{candidate} set $\cC(a)$, and the \textit{bulk} set $\cB(a)$ such that $\cA(a) \cup \cC(a) \cup \cB(a) = \cK$. Then, sorting the $\pi_{a}$ w.r.t. $k$ in descending order, we obtain a new order of communities $k_1, \dots, k_K$. The active set $\cA(a)$ contains communities whose cumulative distribution $F(k_i)$ is less than a threshold $\tau \in (0,1]$, i.e. $\cA(a) = \{k_i \in \cK | F(k_i) < \tau \}$. The candidate set $\cC(a)$ includes communities which are in the active set of at least one of the neighbors of node $a$, i.e. $\cC(a) = \{k \in \cK \setminus \cA(a) | \exists b \in \cN(a) \text{ s.t. } k \in \cA(b)\}$. The bulk set $\cB(a)$ contains all the remainder, i.e. $\cB(a) = \cK \setminus (\cA(a) \cup \cC(a))$.
Here, we use $\cN(a)$ to denote the neighbors of node $a$.

The rationale behind this split scheme is two fold. First, due to \textit{sparsity}, at each node only a small number of communities will have meaningful probability while a large number of communities will have very low probability $\pi_{ak}$. We want the communities of low probability to belong to the bulk set, to share a single probability $\pi_{a\bka}$, and thus to be updated by one-shot for all $k\in\cB(a)$. We use $\bka$ to represent the representative community of a bulk set. Second, due to the \textit{locality}, neighboring nodes are likely to have a similar distribution over communities (after all, the model only assigns high probability to links when the associated nodes have high probability of sampling the same community). That is, when a neighbor of node $a$ has a community $k$ in its active set, this community may be a good candidate to become active for node $a$ as well. By maintaining a candidate set we allow communities to spread efficiently to neighboring nodes and thus through the network.

\textbf{One-shot update:} for communities $k \in \cB(a)$, we apply the following approximation of the unnormalized probability in Eqn. \eqref{eqn:local_f}
\begin{align}
&f_{ab}^{(y)}(k\in\cB(a)) \approx \tilde{f}_{ab}^{(y)}(\bka) \nonumber \\=&
\pi_{a\bka}\left\{ \bar{\bt}_a^{y}(1-\bar{\bt}_a)^{(1-y)} \bar{\pi}_{b} + \dt^{y}(1-\dt)^{(1-y)} (1 - \bar{\pi}_{b}) \right\}.
\label{eqn:local_tilde_f}
\end{align}
That is, we replace $\pi_{bk}$ and $\bt_k$ in Eqn. \eqref{eqn:local_f} by
$\bar{\pi}_b = \f{1}{m} \sum_{k\in \cB_m(a)} \pi_{bk}$ and $\bar{\bt}_a = \f{1}{m} \sum_{k\in \cB_m(a)} \bt_{k}$ respectively using a random mini-batch $\cB_m(a)$ of size $m$ sampled from $\cB(a)$. As a result, all $k \in \cB(a)$ share a single value $\tilde{f}_{ab}^{(y)}(\bka)$. Therefore, we can efficiently approximate the normalization constant by
\begin{align}
Z_{ab}^{(y)} = &\sum_{k\in\cK}f_{ab}^{(y)}(k) \approx \tilde{Z}_{ab}^{(y)} \nonumber \\ =& |\cB(a)|\tilde{f}_{ab}^{(y)}(\bka) + \sum_{k \notin \cB(a)} f_{ab}^{(y)}(k).
\end{align}
Note that we only sum over $|\cA(a) \cup \cC(a)| + 1$ terms which will be a much smaller size than $|\cB(a)|$.
Now, to compute the gradient efficiently, we apply the stratified sampling\footnote{Note that, to be more efficient under the approximation, we use a sampling method which is different to the method used in the global update.} for $\cV_n$ by sampling $n_1$ nodes $\cV_1$ from the neighbors $\cN(a)$ and $n_0$ nodes $\cV_0$ from non-neighbors $\cV^* \setminus \cN(a)$ such that $\cV_n = \cV_1 \cup \cV_0$. Then, the sum of gradients for $\cV_n$ in Eqn. \eqref{eqn:local_update} is obtained by
\begin{align}
&\f{N}{|\cV_n|}\sum_{b \in \cV_n} g_{ab}(\phi_{ak})
\nonumber \\ \approx &c_1 \sum_{b \in \cV_1} \left(\f{\tilde{f}_{ab}^{(1)}(k)}{\tilde{Z}_{ab}^{(1)}\phi_{ak}} - \f{1}{\phi_a}\right) + c_0 \sum_{b' \in \cV_0} \left( \f{\tilde{f}_{ab'}^{(0)}(k)}{\tilde{Z}_{ab'}^{(0)}\phi_{ak}} - \f{1}{\phi_a}\right).
\label{eqn:sum_grad}
\end{align}
Here, we set $c_1 = |\cN(a)|/n_1$ and $c_0 = (N - |\cN(a)|)/n_0$ to ensure the unbiasedness of the gradient under stratified sampling. Again, it is important to note that all states in $\cB(a)$ share a single current value $\phi_{a\bka}$ and also the same update equation of Eqn. \eqref{eqn:sum_grad}. Thus for $\cB(a)$ we compute Eqn. \eqref{eqn:sum_grad} only once and update all of them in one-shot. The computation cost becomes $\cO(|\cA(a) \cup \cC(a)||\cV_n|)$ per node in $\cE_{n_t}$ which we expect to be efficient because $|\cA(a) \cup \cC(a)| \ll K$ due to sparsity. For $k \notin \cB(a)$, we simply replace $\tilde{f}_{ab}^{(y)}(k)$ in Eqn. \eqref{eqn:sum_grad} by $f_{ab}^{(y)}(k)$ in Eqn. \eqref{eqn:local_f}, and update individually.

\textbf{Promotion and demotion:} after updating all $|\cA(a) \cup \cC(a)| + 1$ states (communities), we need to update the community split by promoting (e.g. to active or candidate set) or demoting (e.g. to candidate or bulk set) some of the states. To do this, we sort and normalize $\{\phi_{ak}\}$, and obtain the updated cdf $F(k_i)$. Then, we update $\cA(a)$, $\cC(a)$, and $\cB(a)$ based on the threshold $\tau$ and based on the communities that are active in the neighboring nodes. In particular, if the cdf of the bulk state $F(\bka)$ is less than the threshold, we promote some states in the bulk set by a random sampling. In this case, the number of states to promote is equal to int$((\tau - F(\bka_{-1})) / \pi_{a\bka})$. Here $\bka_{-1}$ denotes a state just left to the $\bka$ in the sorted community sequence. We sample a state from the pool of states that are yet not represented anywhere in the graph. The reason is that we wish to avoid creating disconnected communities of nodes, which we believe represent suboptimal local modes in the posterior distribution. Finally, we check which states in $\cB(a)$ can be promoted to $\cC(a)$ by checking the neighboring nodes. We provide the pseudo code of the above algorithm in the Algorithm \ref{alg}.

\begin{algorithm}[t]
\caption{Pseudo-code for each sampling iteration $t$}\label{alg}
\begin{algorithmic}[1]
\STATE Sample a mini-batch $\cE$ of $n_t$ node pairs from $\cE^*$
\FOR {each node $a$ in $\cE$}
	\STATE Sample a mini-batch of nodes $\cV_n(a) = \cV_1(a) \cup \cV_0(a)$ from $\cV^*$
	\STATE Update $\phi_{ak}$ for all $k \in \cA(a) \cup \cC(a)$ using Eqn. \eqref{eqn:sum_grad} and Eqn. \eqref{eqn:local_update}
	\STATE Update $\phi_{a\bka}$ only for the representative bulk state $\bka$ using Eqn. \eqref{eqn:sum_grad} and Eqn. \eqref{eqn:local_update}
    \STATE Sort and normalize to obtain $\{\pi_{ak}\}$ and the cdf $F(k_i)$ for all $|\cA(a) \cup \cC(a)| + 1$ states
    \STATE Promote or demote some states using the updated cdf,  threshold $\tau$, and neighbor information
\ENDFOR
\FOR {$k$ in a random subset of $\cK$}
	\STATE Update $\ta_{k{\{0,1\}}}$ by Eqn. \eqref{eqn:global_update} using $\cE$ and obtain $\bt_k$ from  $\ta_{k{\{0,1\}}}$
\ENDFOR
\end{algorithmic}
\end{algorithm}

\section{Experiments}
\input{experiment.tex}

\section{Conclusion and Future Work}
In this paper we have developed a new scalable MCMC algorithm based on stochastic gradient computations for assortive mixed membership stochatic blockmodels (a-MMSB). The algorithm represents a natural extension of stochastic gradient Riemannian Langevin dynamics (SGRLD) [2] to a-MMSBs. In line with the results reported in [2] for LDA, SGRLD also significantly outperforms its stochastic variational Bayesian counterpart. As was shown in [18], SGRLD algorithms are particularly suited for distributed implementation. We are currently working towards a distributed implementation of our algorithm on a HPC infrastructure allowing us to perform full Bayesian inference on the very large ``Friendster'' network with almost two billion edges. Initial results show that we achieve good perplexity and sample to convergence on the full network.

\section{Acknowledgement}
This work is supported by NSF grant IIS-1216045.

\subsubsection*{References}

[1] P. Gopalan, D. Mimno, S. Gerrish, M. Freedman, and D. Blei.
Scalable Inference of Overlapping Communities. {\it Advances in Neural
Information Processing Systems}, 2012.

[2] S. Patterson and Y. W. Teh. Stochastic Gradient Riemannian Langevin
Dynamics on the Probability Simplex. {\it Advances in Neural Information Processing
Systems}, 2013.

[3] M. Welling and Y. W. Teh. Bayesian Learning via Stochastic Gradient Langevin
Dynamics. {\it International Conference on Machine Learning}, 2011.

[4] D. Blei, A. Ng, M. Jordan, and J. Lafferty. Latent Dirichlet
allocation. {\it Journal of Machine Learning Research}, 2003.

[5] S. Ahn, B. Shahbaba, and M. Welling. Distributed Stochastic Gradient MCMC. {\it International Conference on Machine Learning}, 2014.

[6] S. Ahn, A. Korattikara, and M. Welling. Bayesian Posterior Sampling via Stochastic Gradient Fisher Scoring. {\it International Conference
on Machine Learning}, 2012.

[7] M. Hoffman, D. Blei, C. Wang, and J. Paisley. Stochastic Variational Inference.
 {\it Journal of Machine Learning Research}, 2013.

[8] E. Airoldi, D. Blei, S. Fienberg, and E. Xing. Mixed Membership Stochastic Blockmodels. {\it Journal of Machine Learning Research.}, 2008.

[9] J. Leskovec, J. Kleinberg, and C. Faloutsos. Graph evolution: Densification and shrinking diameters. {\it ACM Trans. Knowl. Discov}, 2007.

[10] RITA. U.S. Air Carrier Trafic Statistics, Bur. Trans. Stats, 2010.

[11] T. Griffiths and M. Steyvers. Finding Scientific Topics. Proceedings of the National academy of Sciences, 2004.

[12] M. Hoffman, D. Blei, and F. Bach, Online Learning for Latent Dirichlet Allocation. {\it Advance in Neural Information Processing systems}, 2010.

[13] M. Wainwright and M. Jordan. Graphical Models, Exponential Families, and Variational Inference. {\it Foundations and Trends in Machine Learning}, 2008.

[14] R. Neal. Probabilistic inference using Markov chain Monte Carlo methods. {\it Technical report CRG-TR  93-1 Department of Computer Science, Universit of Toronto}, 1993.

[15] M. Girolami, and B. Calderhead. Riemann manifold langevin and hamiltonian monte carlo methods. {\it Journal of the Royal Statistical Society: Series B (Statistical Methodology)}, 2011.

[16] M. Jordan, Z. Ghahramani, T. Jaakkola, and L. Saul. An introduction to variational methods for graphical models. Machine learning, 1999.

[17] R. M. Neal. MCMC using Hamiltonian dynamics. In
Brooks, S., Gelman, A., Jones, G., and Meng, X.-L.
(eds.), {\it Handbook of Markov Chain Monte Carlo}. Chapman \& Hall /CRC Press, 2010

[18] Stanford Large Network Dataset Collection, http://snap.stanford.edu/data/

[19] S. Ahn, A. Korattikara, N. Liu, S. Rajan, and M. Welling, Large-Scale Distributed Bayesian Matrix Factorization
using Stochastic Gradient MCMC, {\it ACM SIGKDD Conference on Knowledge Discovery and Data Mining}, 2015.

\end{document}

%% file: experiment.tex
We evaluate the efficiency and accuracy of our algorithm on five datasets [18]: Synthetic [1], US-AIR, NETSCIENCE, RELATIVITY, and HEP-PH, summarized in Table \ref{tab:data}. (The last column is the percentage of link edges among all possible edges.) We compare four algorithms. As exact batch-mode MCMC methods, we use collapsed Gibbs sampling (CGS) and Langevin Monte Carlo (LMC). We also compare to SVI [1] as a state-of-the-art method in variational Bayes. Finally, two of our algorithms are tested, one with and the other without the approximation for large communities. We call these SGMC and SGMC-M, respectively.

\begin{table}
\center
\vspace{-9mm}
\caption{Datasets}
\hspace{-2mm}
\begin{tabular}{@{}lll@{}}
\toprule
Name 		& \# of nodes & \% \\ \midrule
Synthetic   & $75$        & $30$   \\
US-AIR     	& $1.1$k      & $1.2$  \\
NETSCIENCE  & $1.6$K      & $0.3$  \\
RELATIVITY  & $5.2$K      & $0.05$ \\
HEP_PH    	& $12$k       & $0.16$ \\\bottomrule
\label{tab:data}
\end{tabular}
\vspace{-7mm}
\end{table}


We used $\alpha=1/K$ and $\eta=1$ for all of the models and for all experiments unless otherwise stated. For the stepsize annealing schedule we used $\eps_t = (\tau_0+t)^{-\kappa}$ with $\kappa=0.5$ and $\tau_0=1024$ [2]. For the stratified sampling of the global update in SVI and SGMCs, we used $m$ such that the size of non-link edges $N/m$ to be $30 < N/m < 100$. And for the mini-batch size of the stratified sampling of the local update in SGMCs, we used $20$ samples with $10$ from neighbors and $10$ from non-neighboring nodes. For SGMC-M, we used the threshold $\tau=0.9$ by default unless otherwise stated. Also, for the held-out test set, we used 1\% of the total links and non-links.

As the performance metric, we use perplexity which is defined as exponential of the negative average log-likelihood of the data. Given a collection of $T$ samples of the model parameters $\{\bt_t\}$ and $\{\pi_t\}$, the averaged perplexity on the held-out test set $\cE_h$ is
\begin{align}
&\text{perp}_{\text{avg}}(\cE_h|\{\beta_t\},\{\pi_t\})\nonumber \\=&
\exp\left(-\frac{\sum_{(a,b)\in \cE_h}^{}\log \{(1/T)\sm{t}{T}p(y_{ab}|\beta_t,\pi_t)\}}{|\cE_{h}|}\right)
\end{align}

\vspace{-0.3cm}
\subsection{Results}
\begin{figure*}
\centering
    \begin{subfigure}[b]{0.3\textwidth}
   \includegraphics[width=\textwidth]{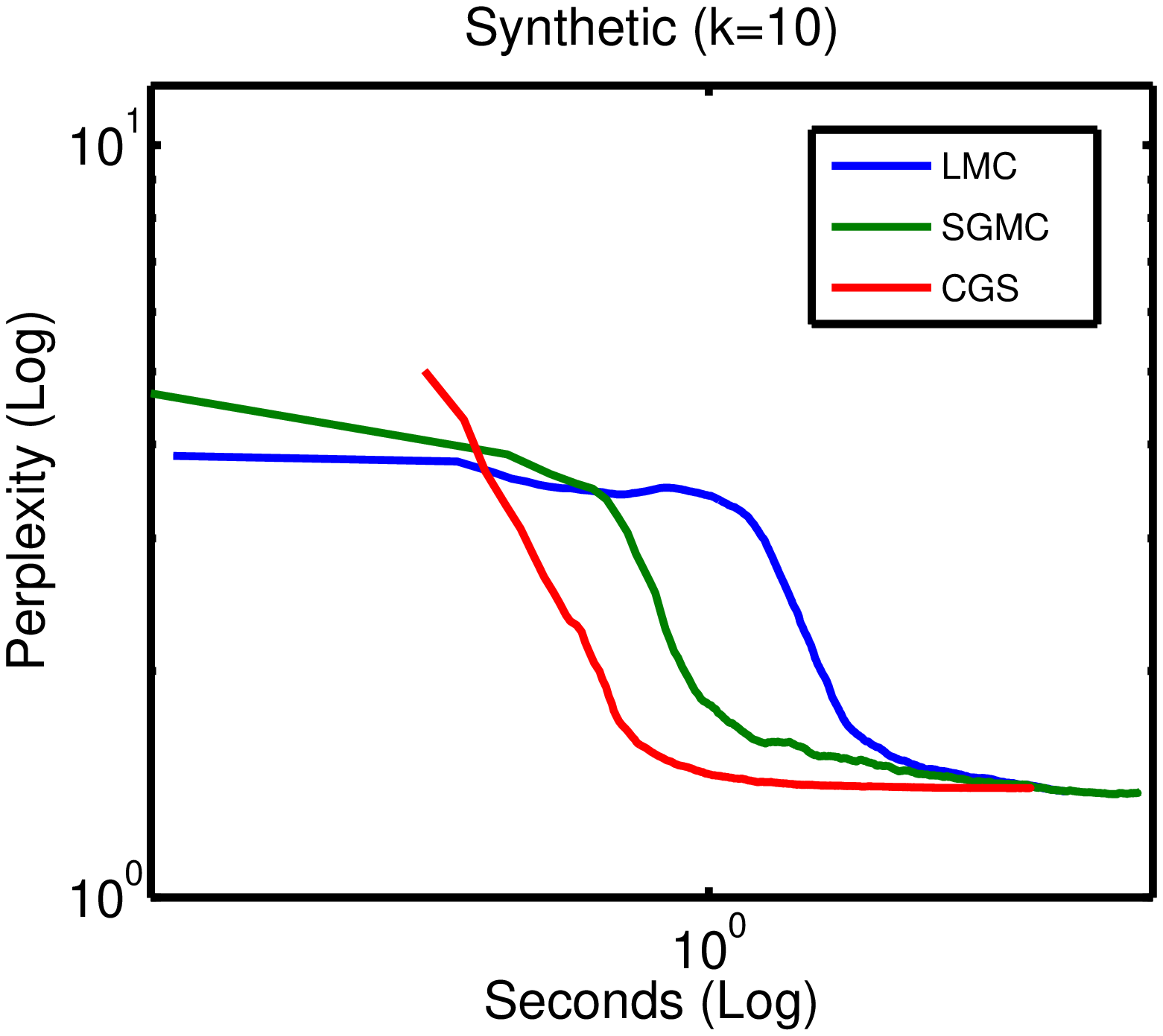}
   \caption{}
   \label{fig:comp_exact_a}
   \end{subfigure}%
    \begin{subfigure}[b]{0.3\textwidth}
   \includegraphics[width=\textwidth]{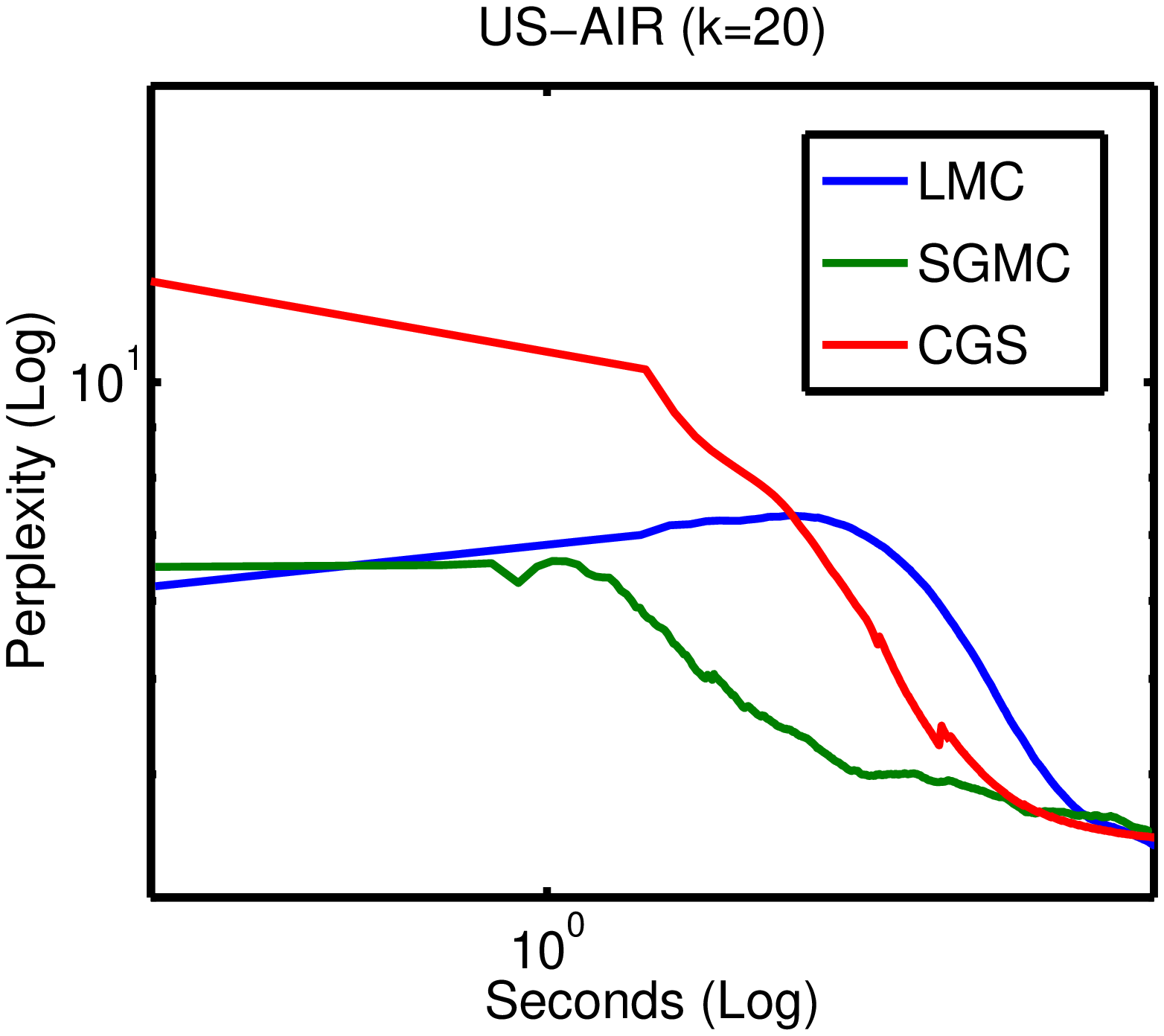}
   \caption{}
   \label{fig:comp_exact_b}
   \end{subfigure}%
      \caption{Convergence of perplexity on (a) Synthetic and (b) US-AIR datasets}\label{fig:compare}
\end{figure*}

\textbf{Comparison to exact batch MCMC:} We first show the accuracy of our algorithm in
comparison to exact batch-mode MCMC
algorithms (CGS and LMC). For this, we use two relatively small datasets, Synthetic and US-AIR, due to the slow speed of the batch algorithms.
The results are shown in Fig. \ref{fig:comp_exact_a} and Fig. \ref{fig:comp_exact_b}.

\begin{figure*}
\centering
    \begin{subfigure}[b]{0.3\textwidth}
   \includegraphics[width=\textwidth]{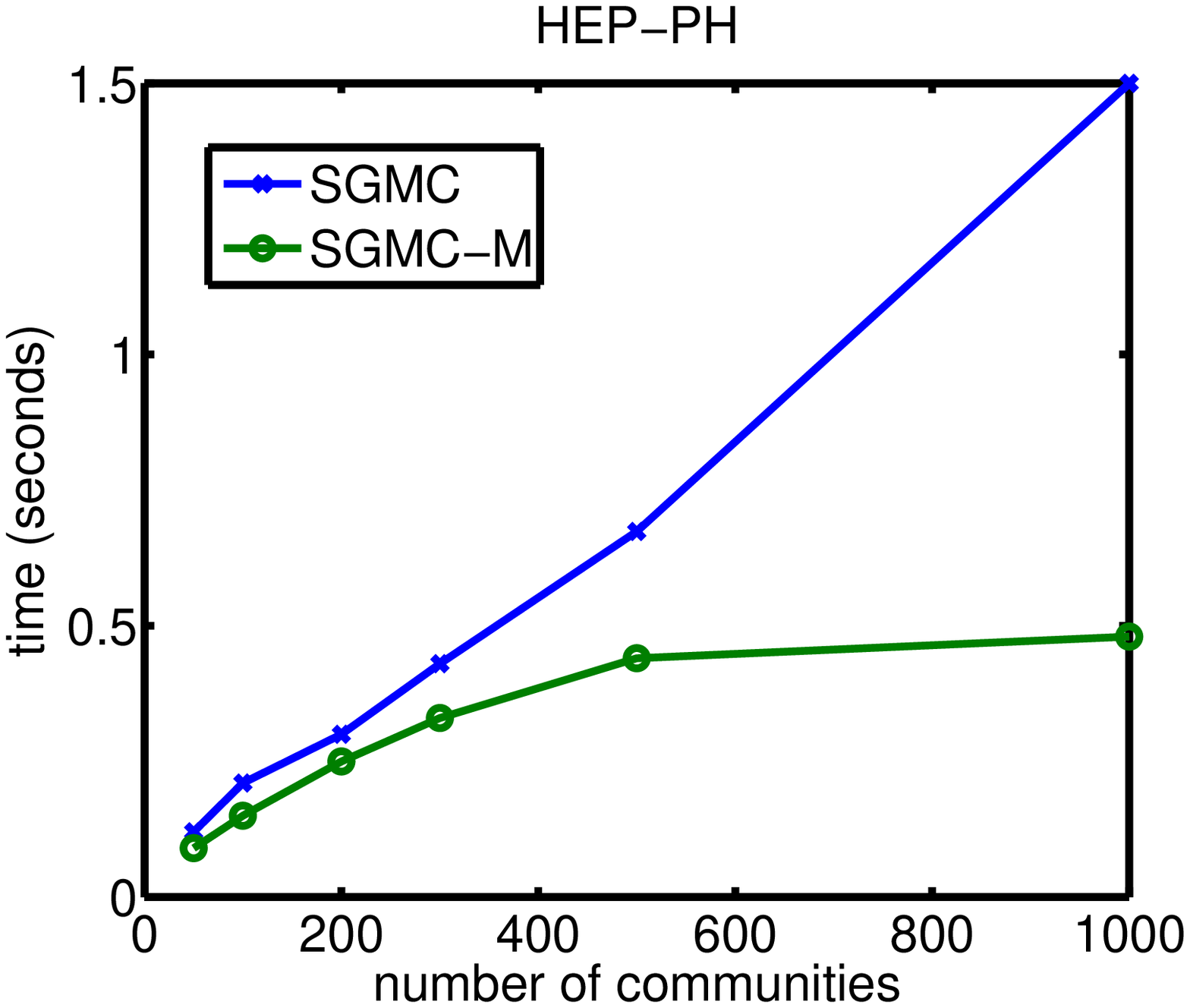}
   \caption{}
   \label{fig:iter_time_hepph}
   \end{subfigure}%
    \begin{subfigure}[b]{0.3\textwidth}
   \includegraphics[width=\textwidth]{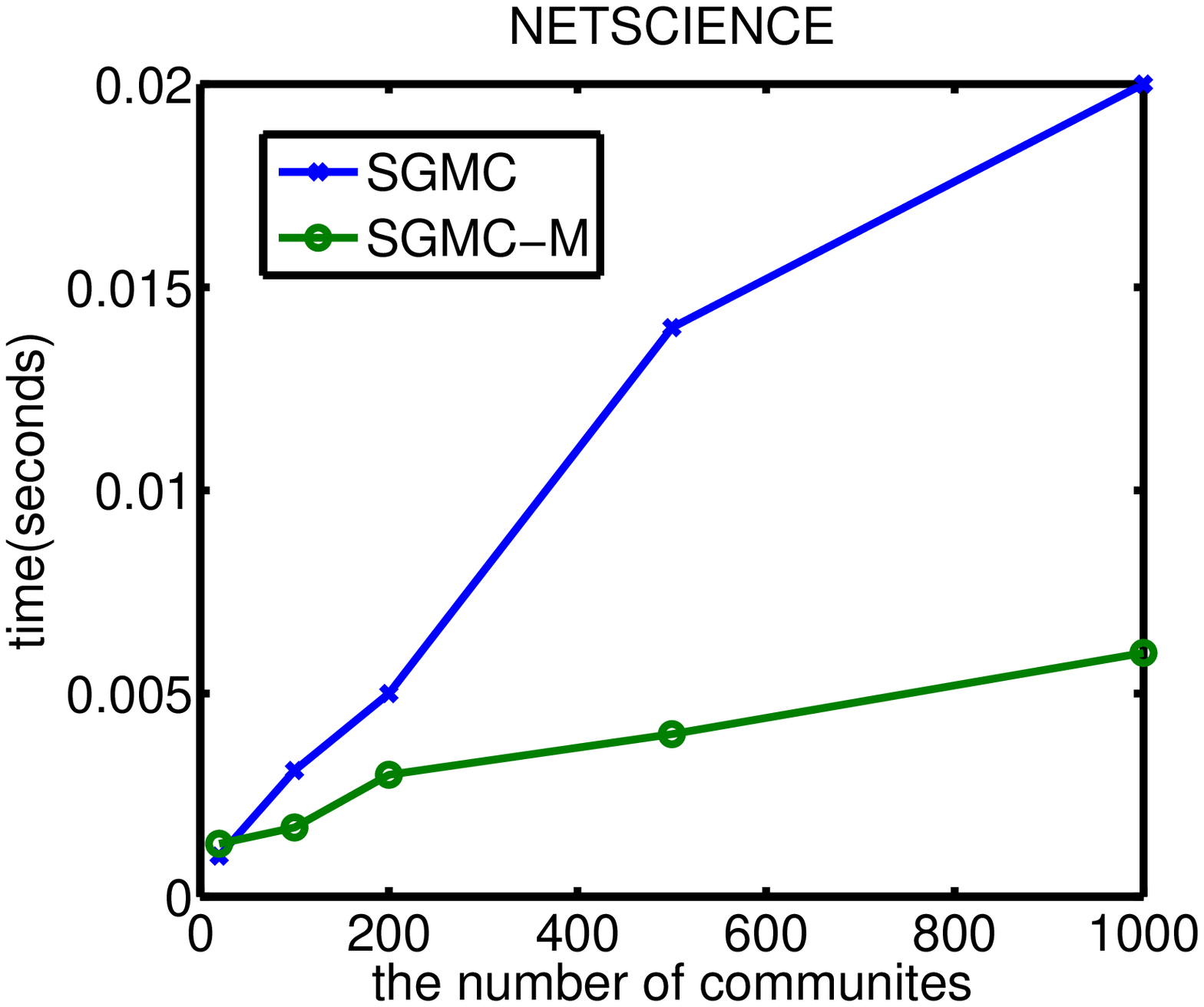}
   \caption{}
   \label{fig:iter_time_netscience}
   \end{subfigure}%
   \begin{subfigure}[b]{0.3\textwidth}
   \includegraphics[width=\textwidth]{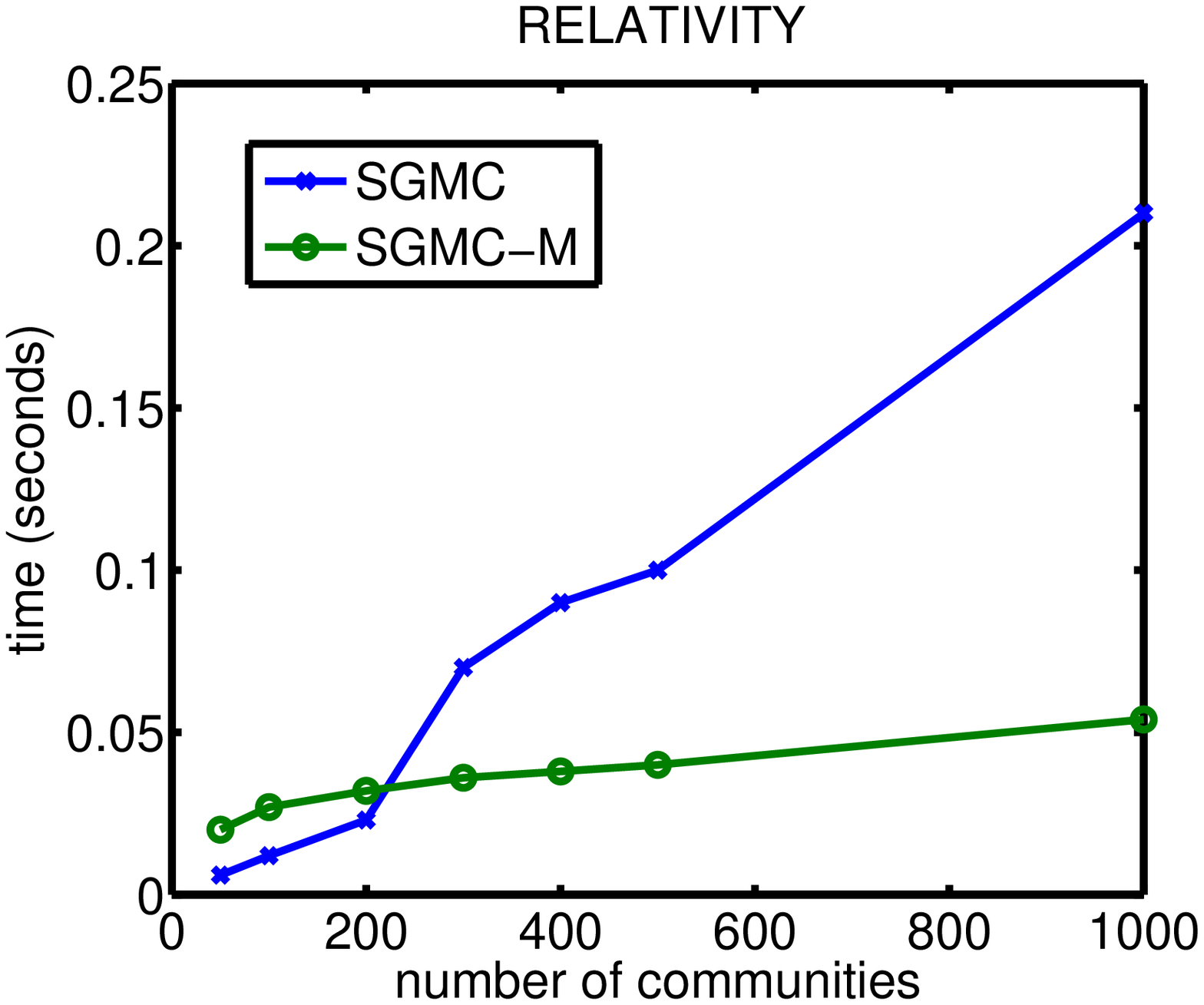}
   \caption{}
   \label{fig:iter_time_relativity}
   \end{subfigure}%
      \caption{(a) Wall-clock time per iteration over increasing community sizes, on (a) HEP-PH, (b) NETSCIENCE and (c) RELATIVITY datasets}\label{fig:time_per_iteration}
\end{figure*}

As expected, for the smaller dataset (Synthetic) in Fig. \ref{fig:comp_exact_a}, we see that CGS converges very fast. However, it is interesting to observe that our stochastic gradient sampler (SGMC) using fixed step-size converges to the same level of accuracy in comparable time, whereas LMC converges much slower than both the collapsed Gibbs sampler and our algorithm due to its full gradient computation and the Metropolis-Hastings accept-reject step. As we move to a larger network (US-AIR) in Fig. \ref{fig:comp_exact_b}, we begin to see that our stochastic gradient sampler outperforms in speed the collapsed Gibbs sampler as well as the Langevin Monte Carlo. It is interesting to see that the approximation error of our algorithm due to the finite step size and the absence of accept-reject tests is negligible compared to the perplexity of the exact MCMC.


\textbf{Effect of our approximation for large communities:}
In Fig. \ref{fig:iter_time_hepph} and Fig. \ref{fig:iter_time_netscience}, Fig.\ref{fig:iter_time_relativity} on three large datasets, HEP-PH, NETSCIENCE and RELATIVITY, we show the speed-up effect of our approximate method (SGMC-M) compared to the SGMC without the approximation. Here we measure the time per iteration for various community size $K = [30,50,100,200,300,500,1000]$ and set the threshold to $\tau = 0.9$. As shown, we can see that the approximate method SGMC-M only slightly increases the wall-clock time per iteration even if the community size increases. However, without the approximation (SGMC), the time per iteration increases linearly w.r.t. the community size. In fact, we can obtain more time savings as the community size increases further because the level of sparsity, i.e. the number of communities for which each node has non-negligible probability of participation, does not change much when we increase $K$.

\begin{figure*}
\centering
    \begin{subfigure}[b]{0.28\textwidth}
   \includegraphics[width=\textwidth]{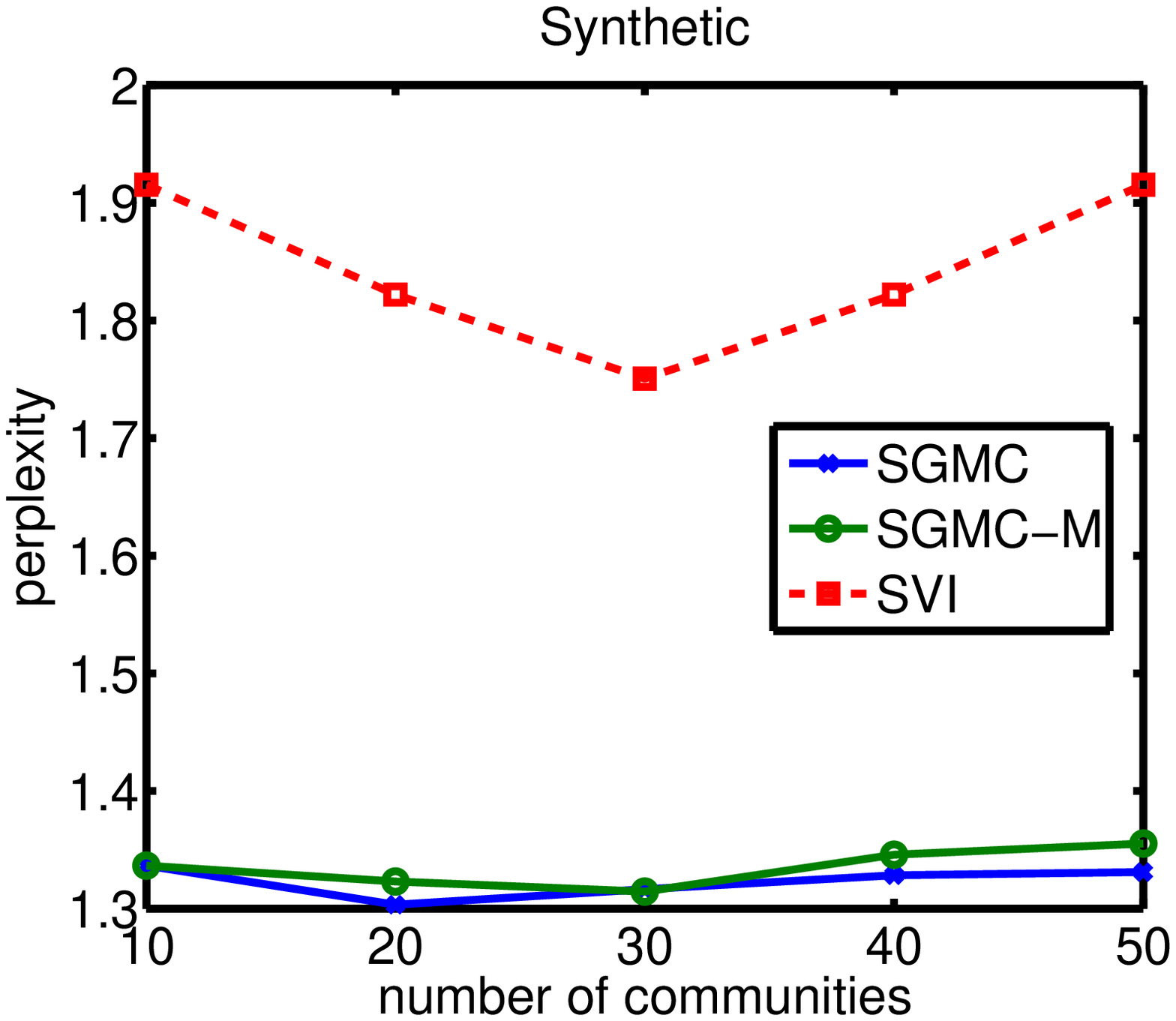}
   \caption{}
   \label{fig:ppx_k_tiny}
   \end{subfigure}%
    \begin{subfigure}[b]{0.28\textwidth}
   \includegraphics[width=\textwidth]{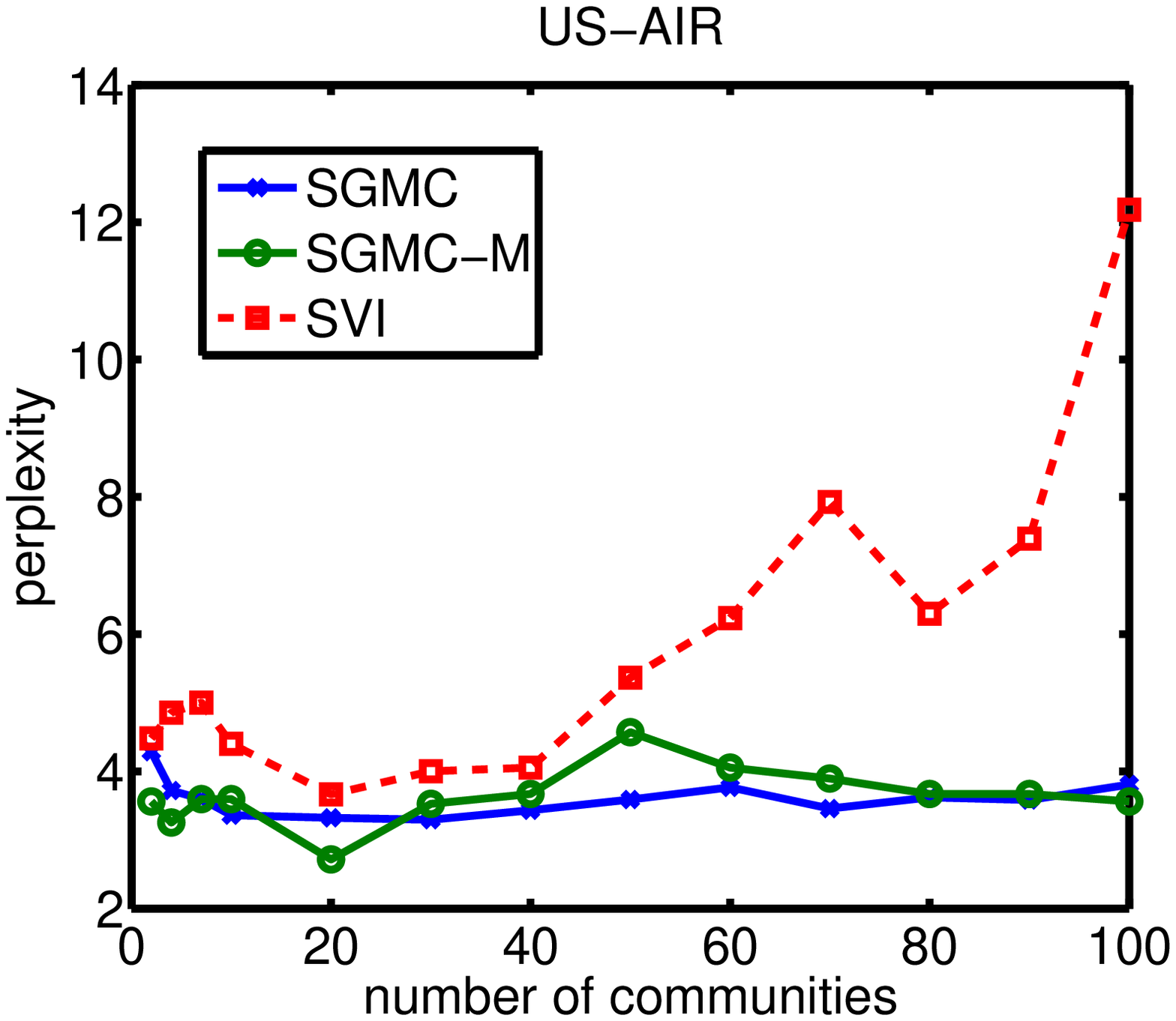}
   \caption{}
   \label{fig:ppx_k_usair}
   \end{subfigure}%
   \begin{subfigure}[b]{0.28\textwidth}
   \includegraphics[width=\textwidth]{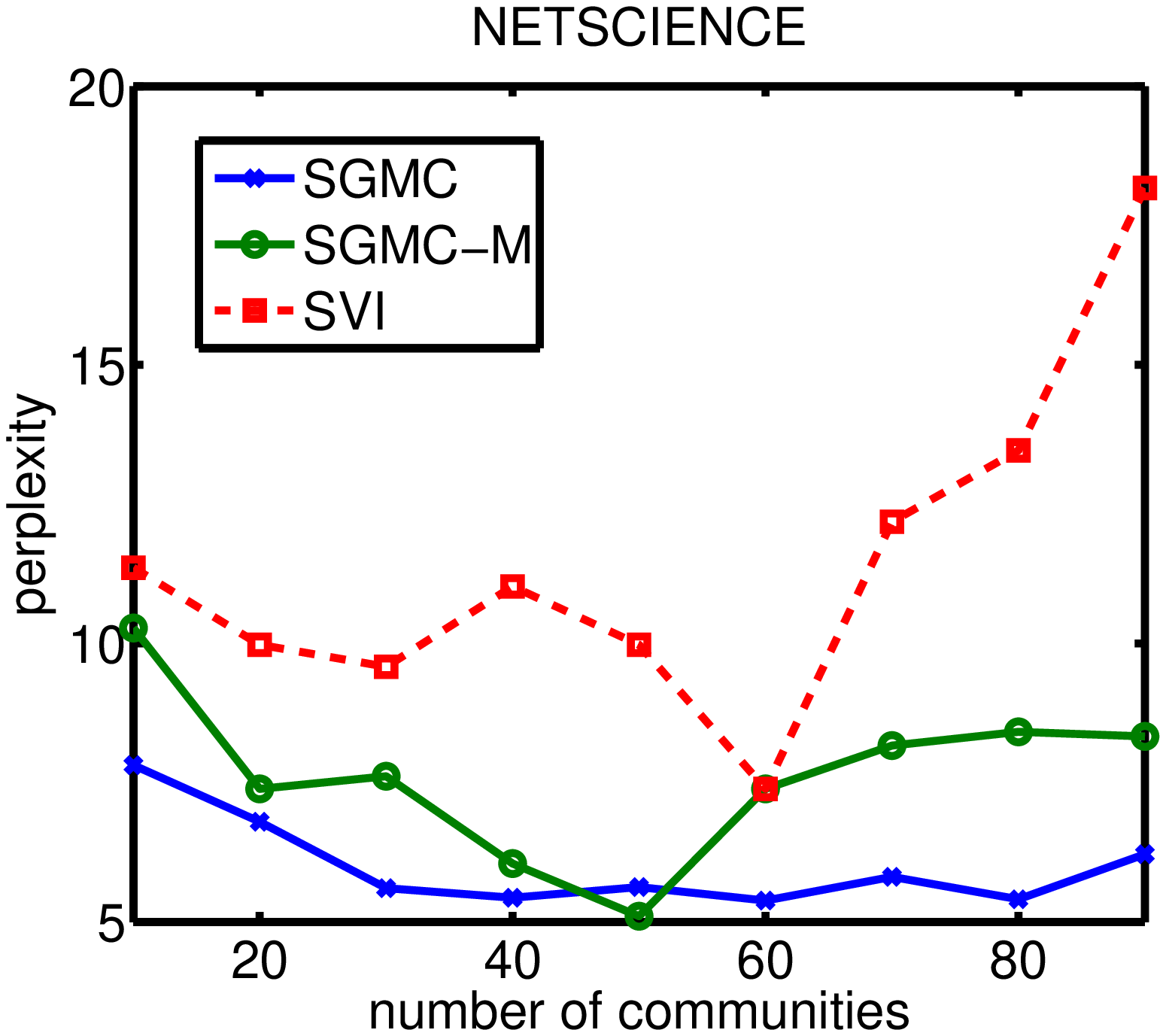}
   \caption{}
   \label{fig:ppx_k_netscience}
   \end{subfigure}%
      \caption{Converged perplexity for various community sizes on (a) Synthetic , (b) US-AIR and (c) NETSCIENCE datasets}\label{fig:ppx_k}
\end{figure*}

Furthermore, it is interesting to see in Fig. \ref{fig:ppx_k_tiny}, Fig. \ref{fig:ppx_k_usair} and Fig. \ref{fig:ppx_k_netscience} that we do not lose much accuracy despite the approximation. In particular, for Fig. \ref{fig:ppx_k_usair}, the SGMC-M performs as good as the SGMC. Although for Fig. \ref{fig:ppx_k_netscience} the SGMC-M performs worse than the SGMC, it still outperforms the SVI. Note that the results are based on converged perplexity which SGMC-M will reach much faster. The figures also reveal some interesting facts. First, the predictive accuracy is dominated by SGMC for all choices of $K$. Second, the curve of SVI has a V-shape indicating that the optimal value for $K$ is in between the minimum and maximum value of $K$ we tested.  However, for SGMC the accuracy remains relatively stable as we increase $K$, making it less sensitive to the choice of this hyperparameter.


\begin{figure*}
\centering
    \begin{subfigure}[b]{0.23\textwidth}
   \includegraphics[width=\textwidth]{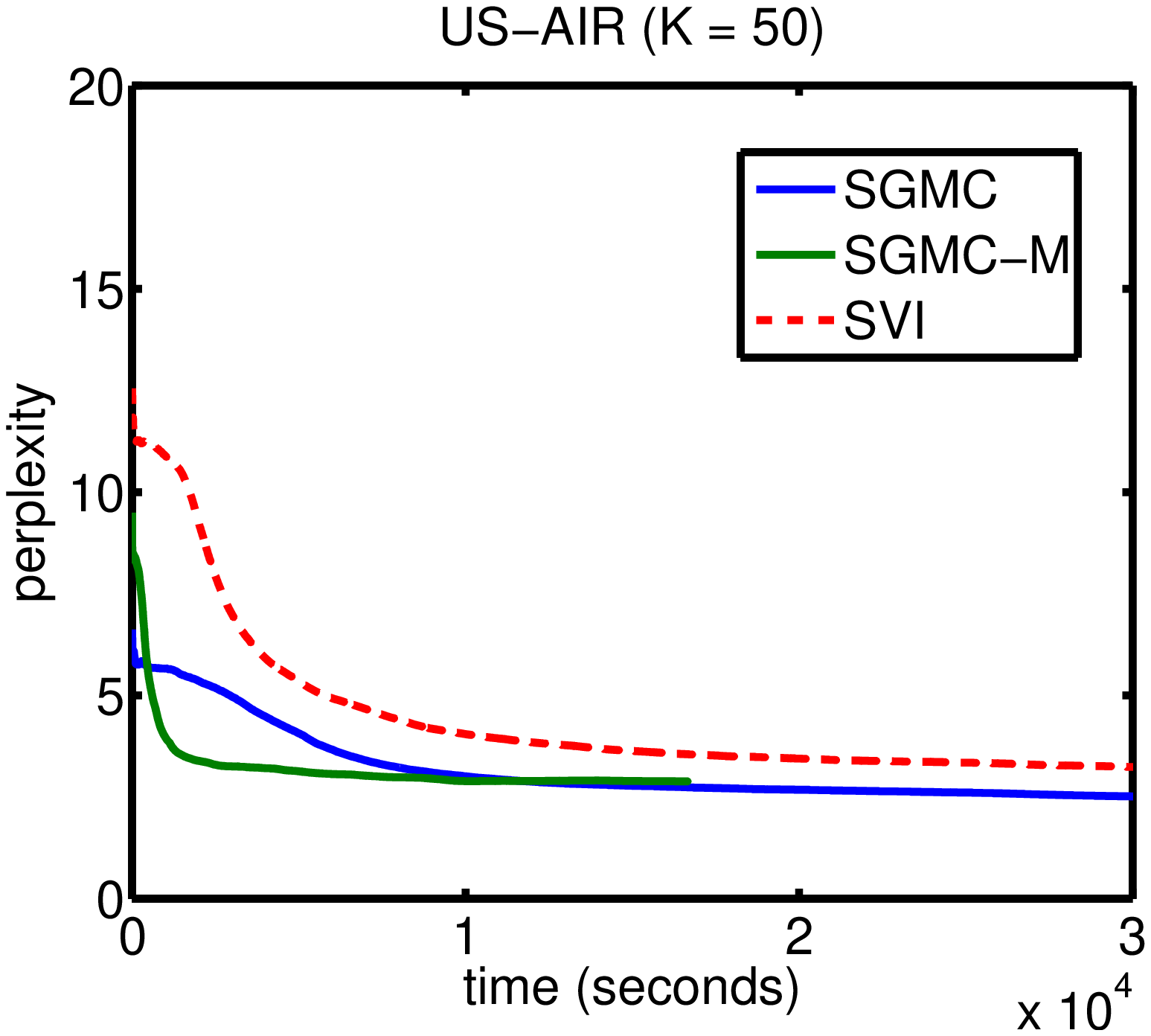}
   \caption{}
   \label{fig:ppx_time_usair_50}
   \end{subfigure}%
    \begin{subfigure}[b]{0.23\textwidth}
   \includegraphics[width=\textwidth]{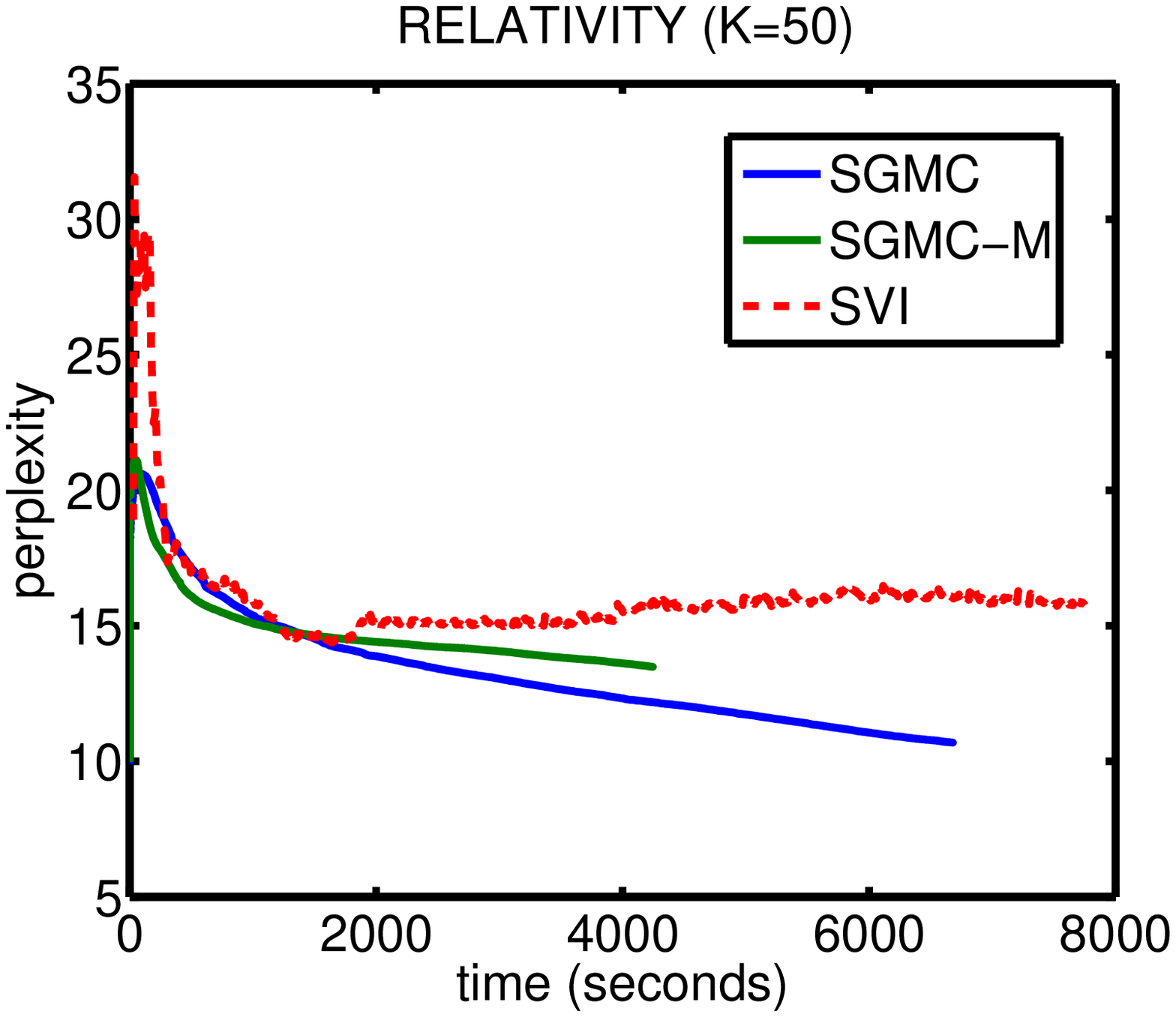}
   \caption{}
   \label{fig:ppx_time_usair_50}
   \end{subfigure}%
   \begin{subfigure}[b]{0.23\textwidth}
   \includegraphics[width=\textwidth]{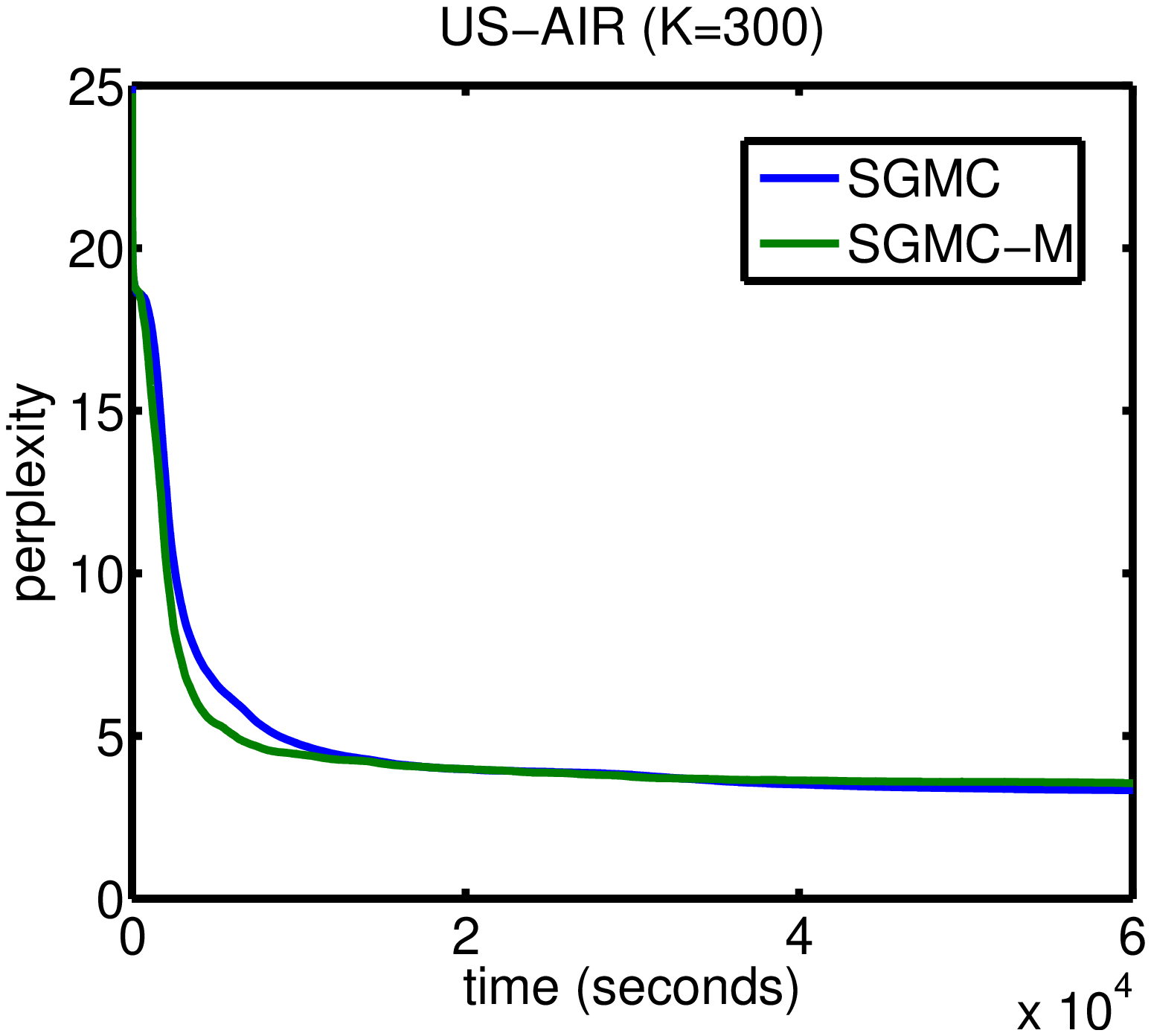}\caption{}\label{fig:ppx_time_usair_300}
   \end{subfigure}%
   \begin{subfigure}[b]{0.23\textwidth}
   \includegraphics[width=\textwidth]{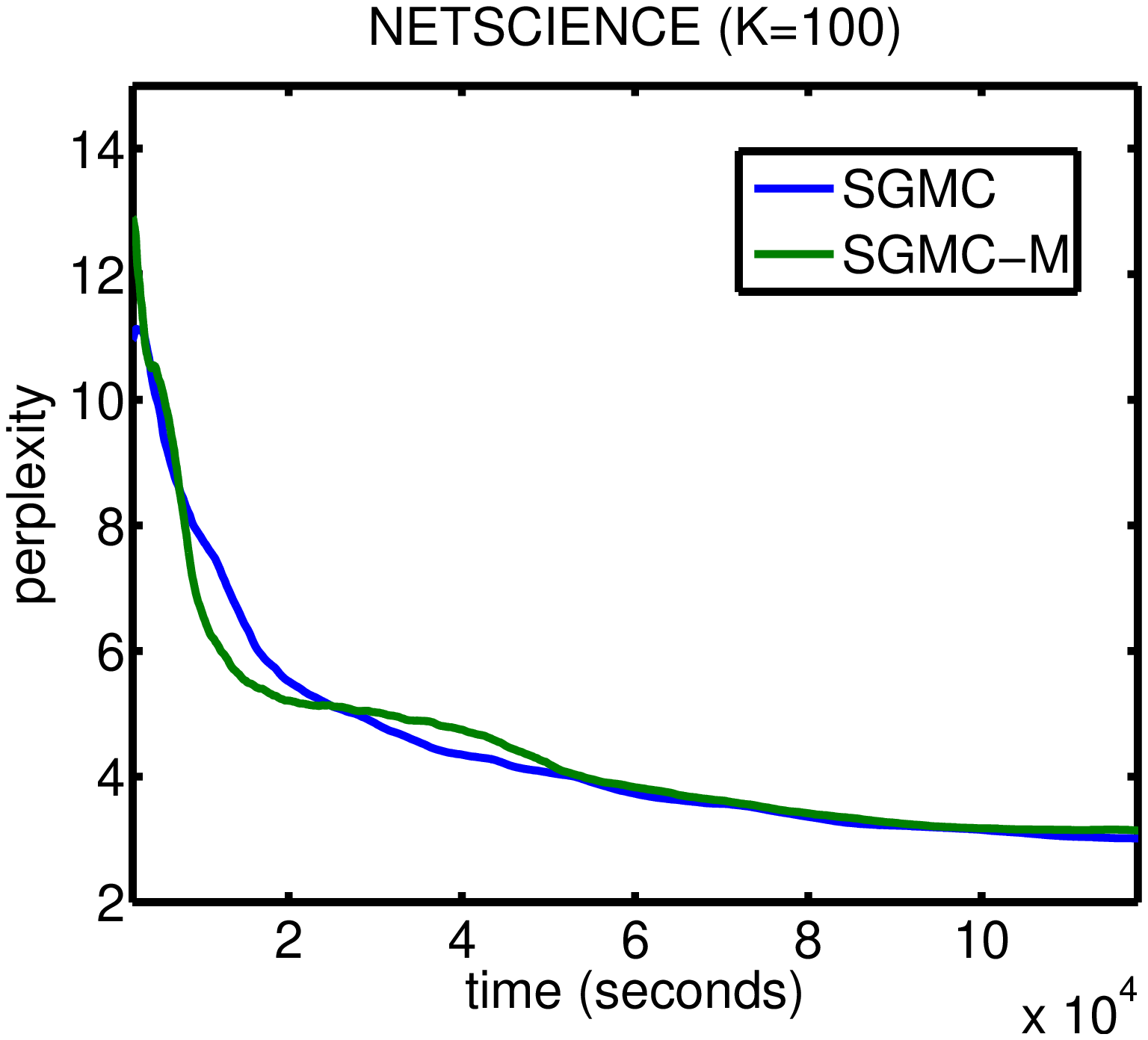}
   \caption{}
   \label{fig:ppx_time_netscience_100}
   \end{subfigure}%
      \caption{The change of perplexity over time on (a) US-AIR (K=50), (b) RELATIVITY (K=50), (c) US-AIR (K=300) and (d) NETSCIENCE (K=100) datasets. For (a) and (b), we compare the performance among three methods: SGMC, SGMC-M and SVI. However, for (c) and (d) we don't include the results for SVI because the perplexity for SVI is hard to compare to our methods.}\label{fig:ppx_time}
\end{figure*}

In Fig. \ref{fig:ppx_time}, we show the convergence of perplexity over wall-clock time on several datasets using SGMC, SGMC-M and SVI. Note that we do not include the SVI method in Fig. \subref{fig:ppx_time_usair_300} and Fig. \ref{fig:ppx_time_netscience_100} where we use a large community size, since the perplexity tends to become very large and thus makes it incomparable to our methods. In general, we have two main observations. First, the approximate method SGMC-M dominates other methods during early stages, but eventually SGMC reaches lower perplexity than SGMC-M. Second, in Fig. \ref{fig:ppx_time_usair_50} and  Fig. \ref{fig:ppx_time_usair_50}, both SGMC and SGMC-M reach much lower perplexity than SVI.

\begin{figure*}[t]
\centering
    \begin{subfigure}[b]{0.3\textwidth}
   \includegraphics[width=\textwidth]{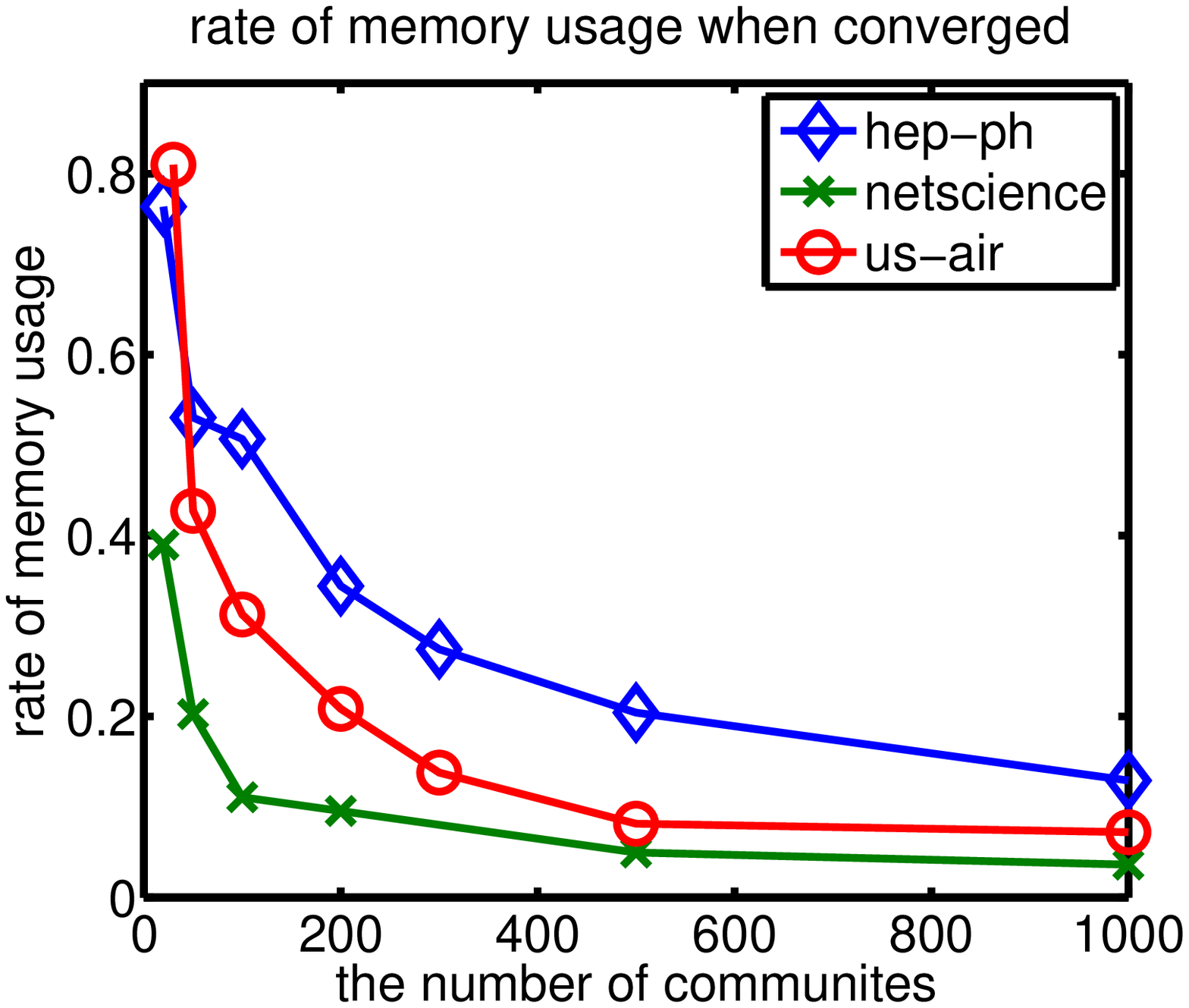}
   \caption{}
   \label{fig:memory_usage}
   \end{subfigure}%
    \begin{subfigure}[b]{0.3\textwidth}
   \includegraphics[width=\textwidth]{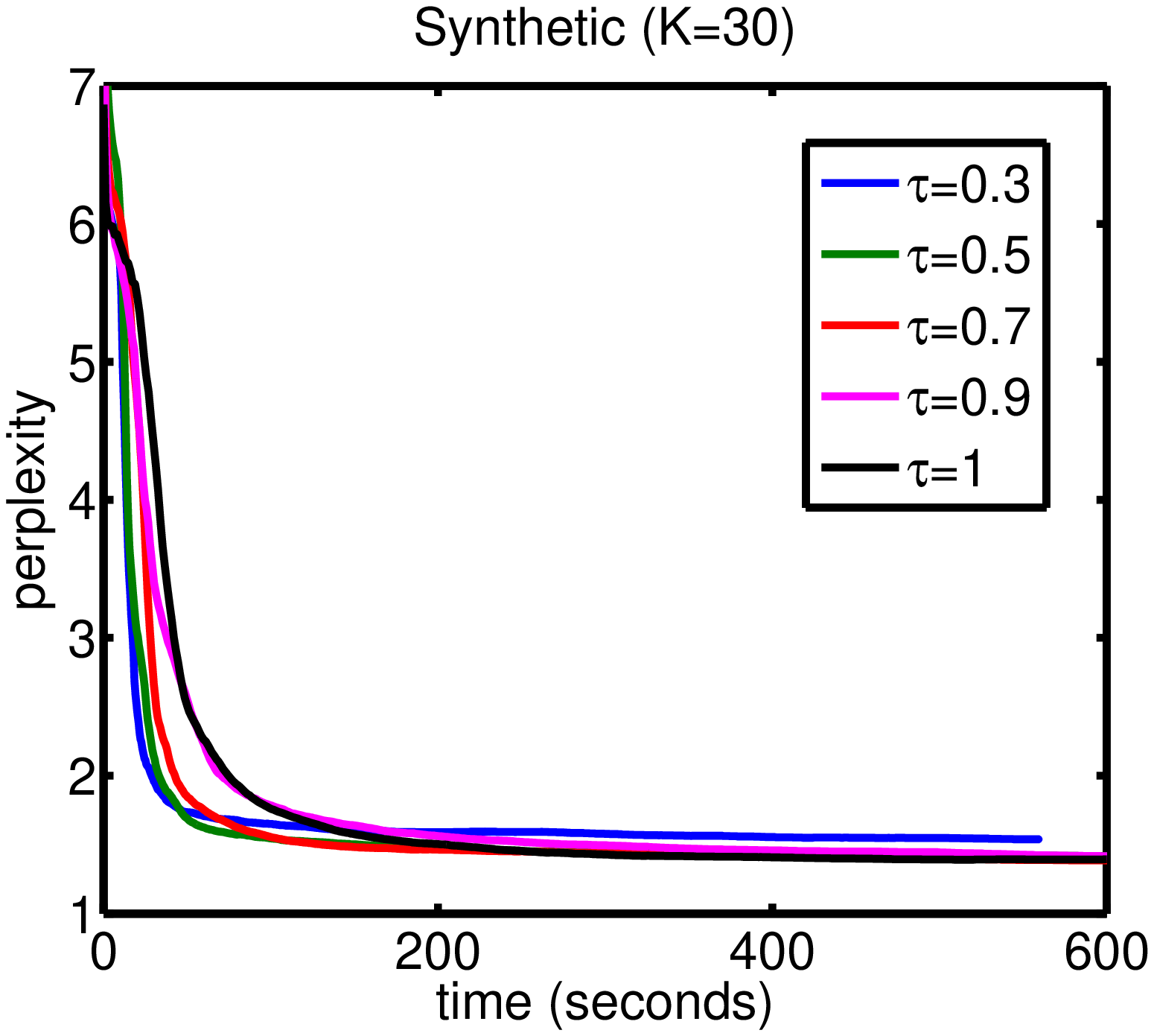}
   \caption{}
   \label{fig:ppx_p_synthetic}
   \end{subfigure}%
   \begin{subfigure}[b]{0.3\textwidth}
   \includegraphics[width=\textwidth]{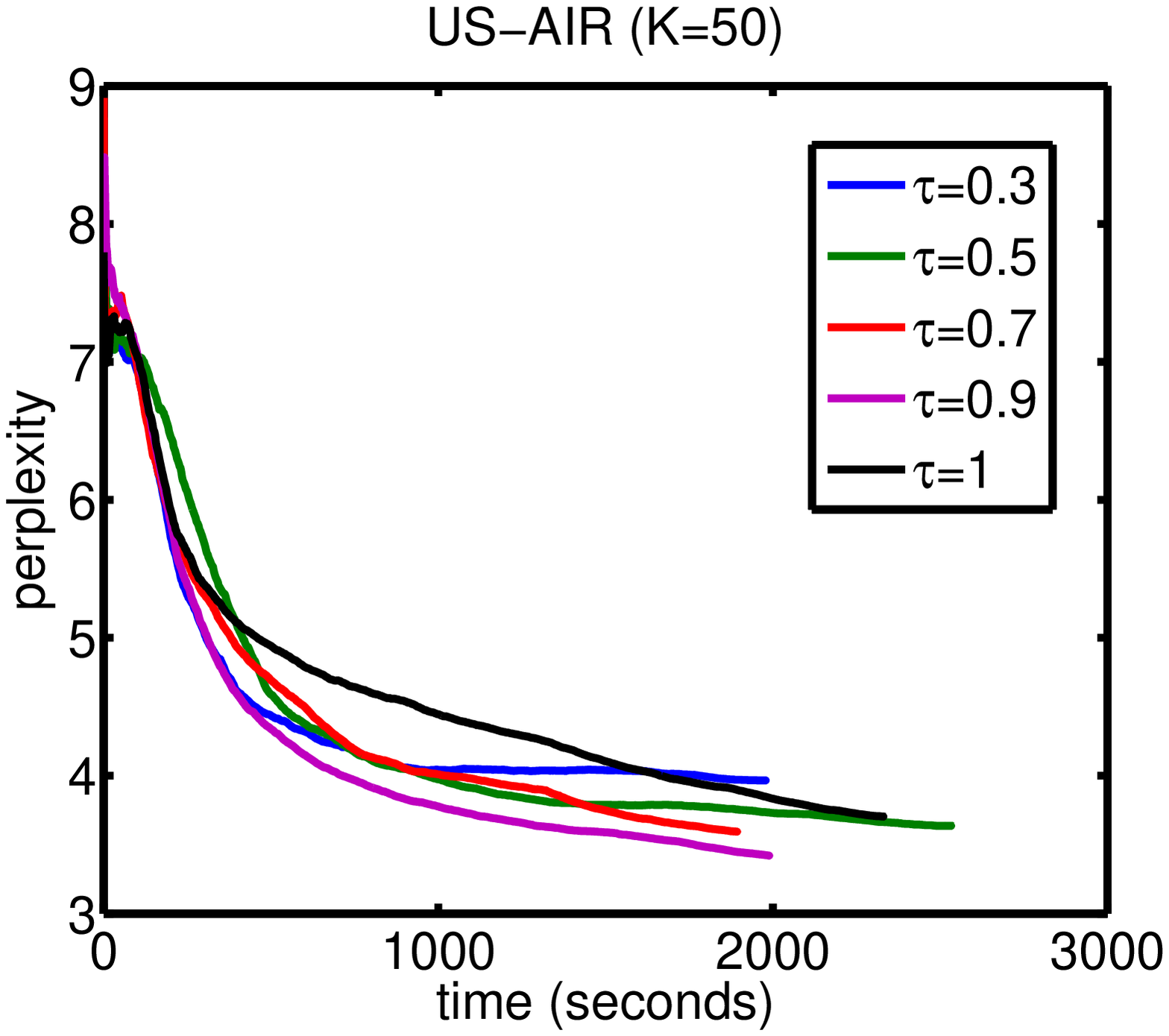}
   \caption{}
   \label{fig:ppx_p_usair}
   \end{subfigure}%
      \caption{(a) Memory usage over different community sizes and datasets. The memory usage is defined as the ratio $(|\cA|+|\cC|+1)/K$. (b) and (c) Convergence of perplexity for various threshold values $\tau$ on Synthetic and US-AIR datasets.}\label{fig:four_fig_1}
\end{figure*}

In Fig. \ref{fig:memory_usage}, we show the efficiency of SGMC-M in terms of memory usage. In this experiment, we set the threshold $\tau = 0.9$ for all datasets. As shown, the memory usage (i.e. $|\cA(a)\cup \cC(a)|/K$) of SGMC-M decreases as the number of communities increases. This is because the sparsity does not change even if we increase the community size $K$. Note that the memory usages of SVI and SGMC are always 100\%. This is a significant feature for large scale networks because without the approximation the space complexity is $\cO(KN)$ where both $K$ and $N$ can be very large [18]. It is also interesting to see that at every node 90\% of the total density is allocated to only about 10\% $\sim$ 20\% of the communities (e.g., for $K=500,1000$).

\begin{figure*}[t]
\centering
    \begin{subfigure}[b]{0.3\textwidth}
   \includegraphics[width=\textwidth]{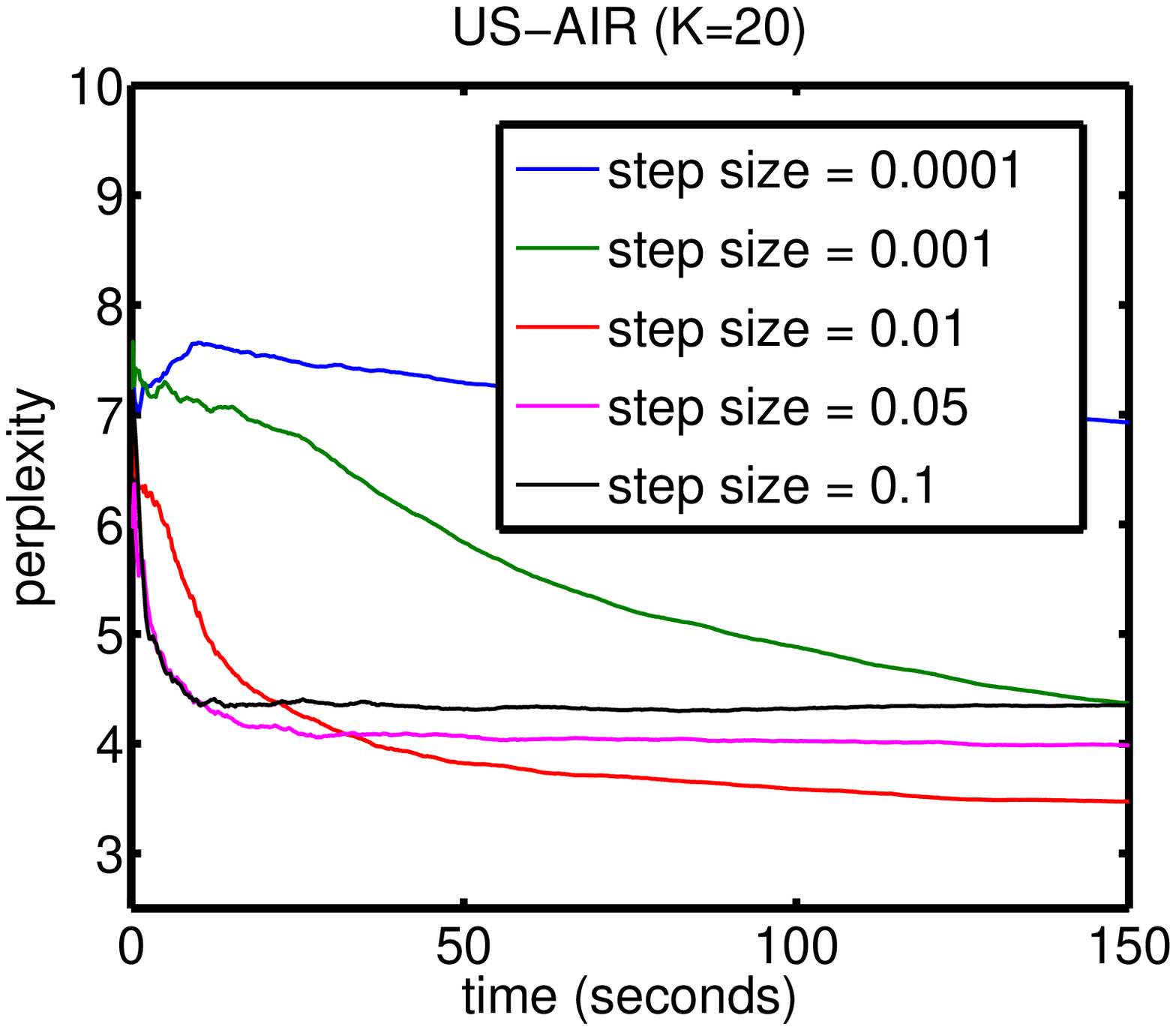}
   \caption{}
   \label{fig:step_size_usair}
   \end{subfigure}%
    \begin{subfigure}[b]{0.3\textwidth}
   \includegraphics[width=\textwidth]{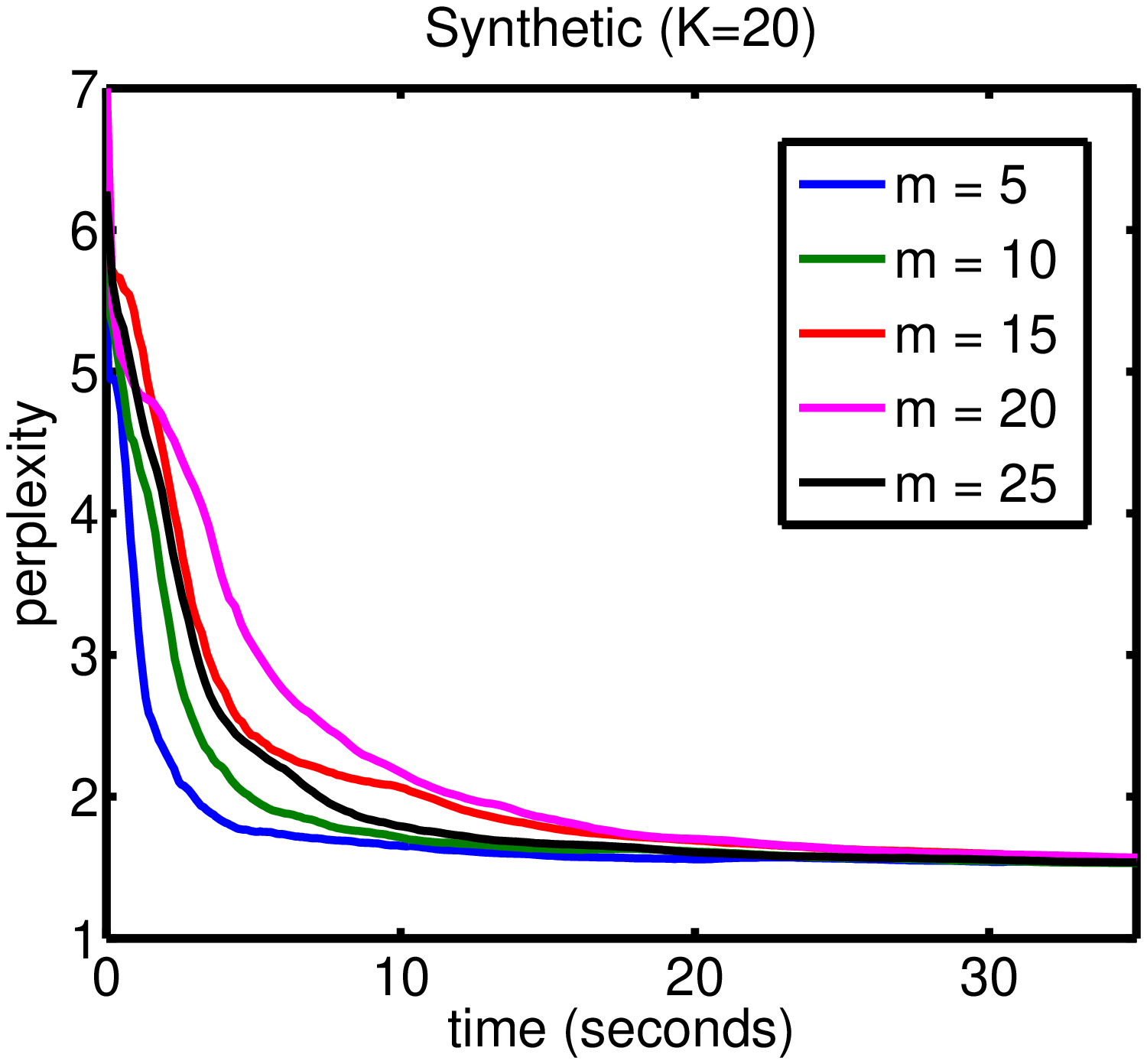}
   \caption{}
   \label{fig:mini_batch_size_synthetic}
   \end{subfigure}%
   \begin{subfigure}[b]{0.3\textwidth}
   \includegraphics[width=\textwidth]{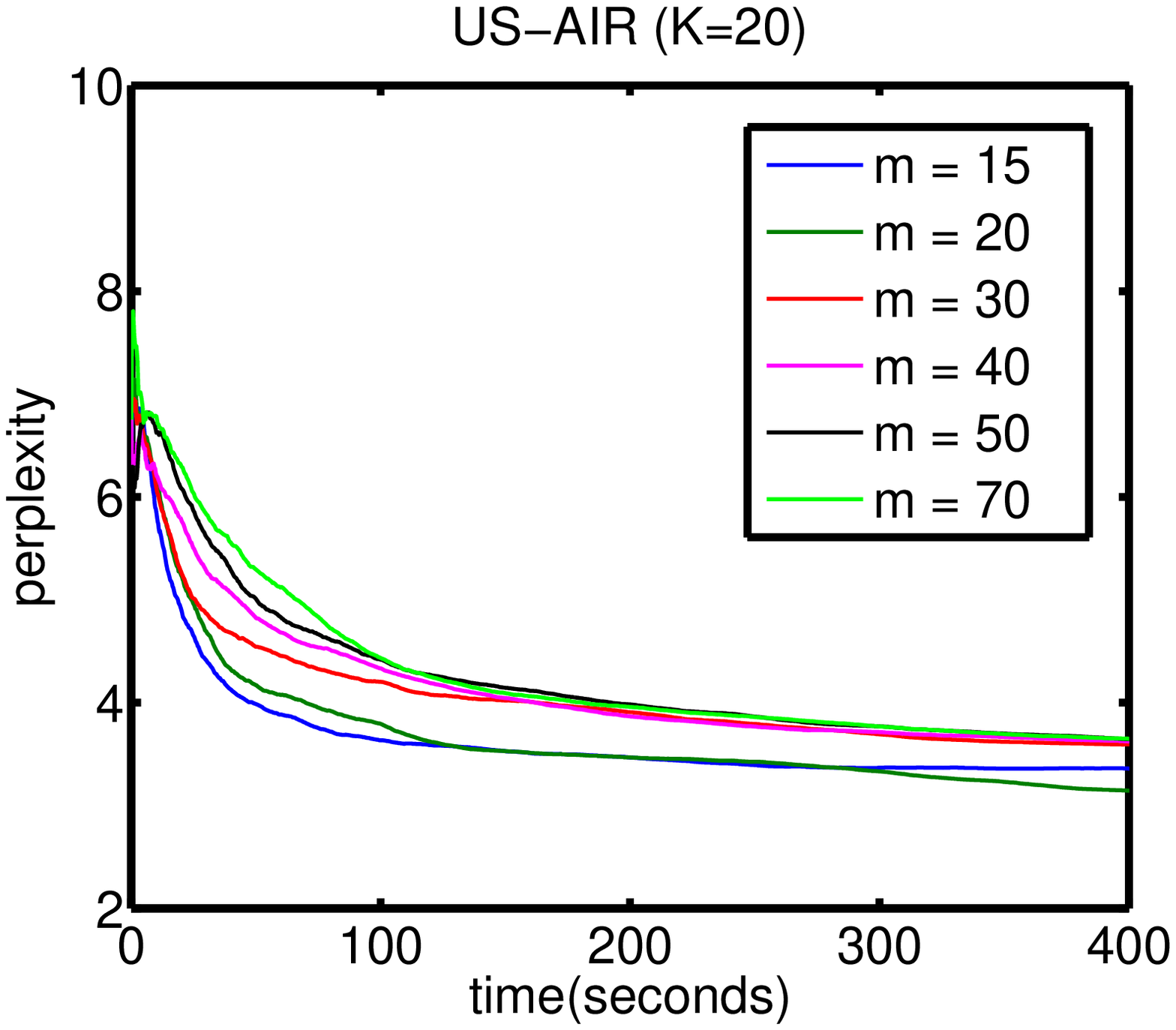}
   \caption{}
   \label{fig:mini_batch_usair}
   \end{subfigure}%
      \caption{(a) Effects of step size. (b) and (c) Effects of mini-batch size. We fix the step size and mini-batch size during the entire training.}\label{fig:four_fig_2}
\end{figure*}

Lastly, we investigate the effect of the threshold $\tau$ and the results are shown in Fig. \ref{fig:ppx_p_synthetic} and Fig. \ref{fig:ppx_p_usair} using Synthetic and US-AIR datasets. As shown, with $\tau=0.9$ and $\tau=1$, we obtain the best result. It is interesting because the memory usage of SGMC-M is only a half of SGMC without the approximation ($\tau=1$). As expected, as we decrease the threshold, smaller communities are represented in the active and candidate set and thus we lose some accuracy while gaining some speed-up.

\paragraph{Effect of step sizes:}
SG-MCMC converges in theory as the step size goes to zero. Although we have used decreasing step sizes in the above experiments, it will be interesting to see how fixed step sizes affect the algorithm, because in practice we cannot decrease the step size to zero. One of the result is shown in Fig. \ref{fig:step_size_usair}. The figure clearly reveals the trade-off between a large step size (leading to a large bias) and a small step size (leading to slow mixing). For the US-AIR dataset a step size $0.01$ seems to work best.

\paragraph{Mini-batch size and computational efficiency:} Depending on the number of edges/nodes sampled in a mini-batch, there is a clear trade-off between the computational cost per update and the variance of an update (i.e. for stochastic
gradients). Our SG-MCMC includes two sampling steps for each iteration, one for edges and the other for nodes (line $1$ and $3$ in Algorithm $1$). Since we are using a ``stratified random node sampling'' strategy, it is important to choose a proper $m$ in order to obtain both fast convergence and low variance. We tested the effects of the parameter $m$. The results are shown in Fig. \ref{fig:mini_batch_size_synthetic} and Fig. \ref{fig:mini_batch_usair}. As we can see, setting $m$ to $5$ achieves the best performance for the synthetic dataset while $m=20$ performs best for the US-AIR data set.\footnote{Note that setting $m$ equals to $5$ for the synthetic data set is roughly equivalent to setting the mini-batch size to $75/m=15$.} In practice, setting $m$ in such a way that the corresponding mini-batch size becomes between $30$ and $100$ results in good performance, i.e. if we have $10$K nodes in total, we could set $m$ to a value between 100-350.\footnote{For the experiments with SVB we choose $m=100$ to make the comparison fair.}. Overall, SG-MCMC takes $O(n^2K)$ computational time\footnote{This complexity is computed by assuming that (i) we use the ``stratified node sampling" and (ii) the average number of links per node is bounded below by $n=N/m$.} for each iteration, where $n=N/m$ and $N$ is the total number of the nodes.